\documentclass[10pt,twocolumn,letterpaper]{article}

\usepackage[accsupp]{axessibility}  % Improves PDF readability for those with disabilities.
\usepackage[pagenumbers]{cvpr} % To force page numbers, e.g. for an arXiv version

\newcommand{\bx}{\mathbf{x}}

\usepackage[pagebackref,breaklinks,colorlinks]{hyperref}
\usepackage{multirow}
\usepackage{cuted}%%\strip
\usepackage{algorithm}
\usepackage{algpseudocode}

\def\paperID{9212} % *** Enter the Paper ID here
\def\confName{CVPR}
\def\confYear{2025}

\title{VideoScene: Distilling Video Diffusion Model to Generate 3D Scenes in One Step}

\author{Hanyang Wang\footnotemark[1] , Fangfu Liu\footnotemark[1] , Jiawei Chi, Yueqi Duan\footnotemark[2]\\
Tsinghua University
}

\begin{document}
\maketitle

\renewcommand{\thefootnote}{\fnsymbol{footnote}}
\footnotetext[1]{Equal contribution. $\dagger$ The corresponding author.}
\renewcommand{\thefootnote}{\arabic{footnote}}

\begin{strip}
    \vspace{-1.5cm}
    \centering
    \includegraphics[width=\textwidth]{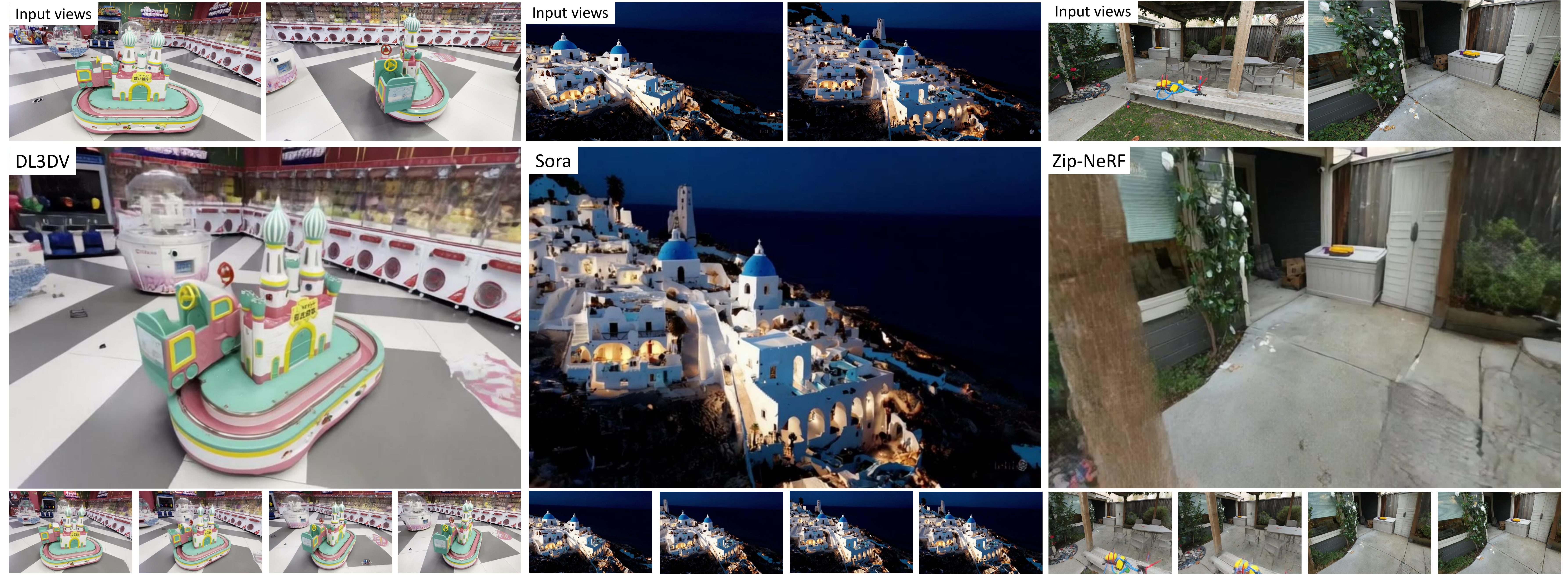}
    \vspace{-0.5cm}
    \captionof{figure}{\textbf{VideoScene} enables one-step video generation of 3D scenes with strong structural consistency from just two input images. The top row shows the input sparse views and the following two rows show the output novel-view video frames.}
    \label{fig:teaser}
\end{strip}

\begin{abstract}
Recovering 3D scenes from sparse views is a challenging task due to its inherent ill-posed problem. Conventional methods have developed specialized solutions (e.g., geometry regularization or feed-forward deterministic model) to mitigate the issue. However, they still suffer from performance degradation by minimal overlap across input views with insufficient visual information. Fortunately, recent video generative models show promise in addressing this challenge as they are capable of generating video clips with plausible 3D structures. Powered by large pretrained video diffusion models, some pioneering research start to explore the potential of video generative prior and create 3D scenes from sparse views. Despite impressive improvements, they are limited by slow inference time and the lack of 3D constraint, leading to inefficiencies and reconstruction artifacts that do not align with real-world geometry structure. In this paper, we propose \textbf{VideoScene} to distill the video diffusion model to generate 3D scenes in one step, aiming to build an efficient and effective tool to bridge the gap from video to 3D. Specifically, we design a 3D-aware leap flow distillation strategy to leap over time-consuming redundant information and train a dynamic denoising policy network to adaptively determine the optimal leap timestep during inference. Extensive experiments demonstrate that our VideoScene achieves faster and superior 3D scene generation results than previous video diffusion models, highlighting its potential as an efficient tool for future video to 3D applications. Project Page: \url{https://hanyang-21.github.io/VideoScene}.
\end{abstract}    
\vspace{-0.5cm}
\section{Introduction}
\label{sec:intro}
The demand for efficient 3D reconstruction is growing rapidly, driven by applications in real-time gaming~\cite{valevski2024diffusion}, autonomous navigation~\cite{adamkiewicz2022vision}, and beyond~\cite{martin2021nerf_in_wild, ye2024dreamreward}. Techniques like NeRF~\cite{mildenhall2021nerf} and 3DGS~\cite{kerbl2023_3dgs} have pioneered high-quality, dense-view reconstruction and demonstrated impressive performance in realistic scene generation. However, these methods typically require numerous professionally captured images, limiting accessibility~\cite{wang2023sparsenerf}. To overcome this, researchers have begun exploring 3D reconstruction from sparse views~\cite{wang2023sparsenerf,yang2023freenerf,yu2024lm-gaussian,charatan2024pixelsplat}, reducing the input requirements to even two casually captured images.

Previous work has developed various specialized methods to tackle the challenges of sparse-view reconstruction, such as geometry regularization techniques~\cite{yang2023freenerf,somraj2023simplenerf} for sparse inputs and feed-forward models~\cite{charatan2024pixelsplat,chen2025mvsplat} trained to create 3D scene from a few images. While these methods produce reasonable scene-level results from nearby viewpoints, they struggle to overcome the under-constrained nature of sparse-view reconstruction, where complex 3D structures have to be inferred from limited visual information~\cite{gao2024cat3d}. The scarcity of views makes it difficult to recover a complete and coherent 3D scene. Thus, there is a need for methods that can both integrate minimal visual information and generate plausible missing details to reconstruct realistic 3D geometry.

Recent advancements in video generative models~\cite{brooks2024sora,xing2025dynamicrafter,hong2022cogvideo,yang2024cogvideox} offer new promise, as these models are capable of generating sequences with plausible 3D structures. Leveraging large, pretrained video diffusion models, early studies~\cite{liu2024reconx,liu20243dgs-enhancer,yu2024viewcrafter,chen2024mvsplat360,sun2024dimensionxcreate3d4d,liu2024physics3d} have explored using video generative priors for 3D scene creation from sparse views. Despite these advances, current models face significant limitations, particularly in two areas: (1) \textit{lengthy inference time} as video diffusion models require multiple denoising steps to generate a high-quality video from a pure noisy input, making them far less efficient than many feedforward methods; (2) \textit{lack of 3D constraints} as these models are trained on 2D video data, focusing on RGB space and temporal consistency rather than stable 3D geometry. As a result, generated videos often lack spatial stability and accurate camera geometry, leading to reconstruction artifacts that hinder their effectiveness for real-world 3D applications. This motivates us to build an efficient and effective tool to bridge the gap from video to 3D.

In this paper, we introduce \textbf{VideoScene}, a novel video distillation framework that optimizes video diffusion models for efficient, one-step 3D scene generation. Our approach tackles the inefficiencies of traditional diffusion steps by reducing redundant information—such as dynamic motions and object interactions—that detract from 3D consistency. We identify that the primary bottleneck stems from the low-information starting point of pure noise in traditional denoising, which makes distillation slow and unstable. To address this, we propose a 3D-aware leap flow distillation strategy to leap over time-consuming denoising stages. Specifically, we first give two input images with corresponding camera poses (can also be estimated via COLMAP~\cite{schonberger2016colmap} or DUSt3R~\cite{wang2024dust3r}). Here, We focus on image pairs for two main reasons. First, monocular 3D reconstruction is inherently ill-posed; using two images enables triangulation between rays from different viewpoints, allowing for a more robust 3D estimation. Second, two-view geometry serves as the fundamental building block for multi-view geometry, which can be extended to full multi-view reconstruction. Then We use a rapid, feed-forward sparse-view 3DGS model~\cite{chen2025mvsplat} to generate a coarse but 3D-consistent scene, rendering frames along an interpolated camera path. This initial 3D-aware video establishes a strong prior that guides subsequent diffusion steps, enabling us to leap over uncertain phases and focus on the deterministic target scene in the consistency distillation~\cite{song2023consistency_model}. Additionally, we develop a dynamic denoising policy network (DDPNet) that learns to adaptively select the optimal leap timestep during inference. This policy maximizes the use of 3D priors, balancing noise addition with information retention, thereby improving efficiency without sacrificing quality. Extensive experiments demonstrate that VideoScene outperforms existing video diffusion methods in both fidelity and speed across a range of real-world datasets~\cite{ling2024dl3dv,brooks2024sora,barron2023zipnerf}. VideoScene shows significant potential as a versatile plug-and-play tool for future applications in 3D scene reconstruction from video generation models as shown in Fig.~\ref{fig:teaser}. In summary, our main contributions are:
\begin{itemize}[leftmargin=*]
    \item We introduce VideoScene, a novel video distillation framework that distills video diffusion models to generate 3D scenes in one step.
    \item We propose a 3D-aware leap flow distillation strategy to leap over low-information steps by providing 3D prior constraints.
    \item We design a dynamic denoising policy network to decide optimal leap timesteps by integrating contextual bandit learning into the distillation process.
    \item Extensive experiments demonstrate that our VideoScene outperforms current methods in both quality and efficiency across diverse real-world datasets.
\end{itemize}
\section{Related Work}
\textbf{Video Generation.}
Efficient and high-quality video generation has become a highly popular topic in recent research. Early methods~\cite{tulyakov2018mocogan, skorokhodov2022stylegan, balaji2019conditional, ge2022long} primarily relied on generative adversarial networks (GANs), which often resulted in low quality and poor generalization to unseen domains~\cite{blattmann2023align}. With the rise of diffusion models in text-to-image generation, recent studies~\cite{blattmann2023svd,ho2022imagen_video,wu2023tune,wang2023modelscope,wei2024dreamvideo,wang2024videocomposer,hong2022cogvideo,yang2024cogvideox} have explored the potential of diffusion-based text-to-video (T2V) generation, achieving promising results. Some methods~\cite{wang2023modelscope,blattmann2023align} focus specifically on generating temporally consistent videos, incorporating temporal modules into 2D UNet architectures and are trained on large-scale datasets. Many works have also explored image-conditioned video synthesis. Image-to-video (I2V) diffusion models, such as DynamiCrafter~\cite{xing2025dynamicrafter}, SparseCtrl~\cite{guo2025sparsectrl}, and PixelDance~\cite{zeng2024pixeldance}, show strong versatility for tasks like video interpolation and transition. As datasets continue to grow, the SORA-like DiT-based (Transformer-based Diffusion Model)~\cite{peebles2023DiT,brooks2024sora} video generation models show clear advancements over earlier UNet-based models~\cite{blattmann2023svd,guo2023animatediff,chen2024videocrafter2}. These models offer enhanced expressive capabilities, improved visual quality, and effective multi-modal integration. With these improvements, they achieve near-production-level performance and demonstrate strong potential for commercial applications. 

\noindent\textbf{Consistency Model.}
Diffusion models~\cite{ho2020ddpm,song2020ddim,song2020score-based,rombach2022ldm} have demonstrated strong performance across various generative tasks. However, they face a speed bottleneck due to the need for a large number of inference steps. To address this, researchers have proposed several solutions, including ODE solvers~\cite{song2020ddim,lu2022dpm-solver}, adaptive step-size solvers~\cite{jolicoeur2021gotta}, neural operators~\cite{zheng2023fast}, and model distillation~\cite{salimans2022progressive,meng2023distillation,sauer2025ADD,kohler2024imagine_flash}. Among these approaches, consistency models~\cite{song2023consistency_model} have shown particular promise. Based on the probability flow ordinary differential equation (PF-ODE), consistency models are trained to map any point in the generation process back to the starting point, or the original clean image. This approach enables one-step image generation without losing the benefits of multi-step iterative sampling, supporting high-quality output. Consistency models can be derived in two ways: either through distillation from a pre-trained diffusion model (\ie, Consistency Distillation) or by training directly from scratch (\ie, Consistency Training). Building on this framework, LCM~\cite{luo2023latent_consistency_model,luo2023lcm-lora} further explores consistency models within latent space to reduce memory consumption and enhance inference efficiency. Subsequent methods~\cite{lu2024simplifying,song2023improved,geng2024cm_made_easy} have also refined these efficiency improvements, achieving impressive results.

\noindent\textbf{Video for 3D Reconstruction.}
In 3D reconstruction, methods like NeRF~\cite{mildenhall2021nerf} and 3DGS~\cite{kerbl2023_3dgs} typically require hundreds of input images for per-scene optimization, which is impractical for casual users in real-world applications. Recent research~\cite{liu2024sherpa3d,liu2024make-your-3d,charatan2024pixelsplat,wewer2024latentsplat,szymanowicz2024splatter_image,chen2025mvsplat,wu2024unique3d} has focused on developing feed-forward models that generate 3D representations directly from only a few input images. pixelSplat~\cite{charatan2024pixelsplat} devised pixel-aligned features for 3DGS reconstruction, using the epipolar transformer to better extract scene features. 
Following that, MVSplat~\cite{chen2025mvsplat} introduces multi-view feature extraction and cost volume construction to capture cross-view feature similarities for depth estimation. The per-view depth maps are predicted and unprojected to 3D to form Gaussian centers, producing high-quality 3D Gaussians in a faster way.
% Following that, MVSplat~\cite{chen2025mvsplat} introduces the cost volume and depth refinements to produce high-quality 3D Gaussians in a faster way. 
While these models can produce highly realistic images from provided viewpoints, they often struggle to render high-quality details in areas not visible from these limited perspectives. To address this challenge, many studies~\cite{liu2024reconx,yu2024viewcrafter,liu20243dgs-enhancer,chen2024mvsplat360,sun2024dimensionxcreate3d4d} have employed large-scale video diffusion models to generate pseudo-dense views from sparse input, aiming to transfer generalization capabilities for the under-constrained sparse-view reconstruction. However, the inefficiency of multi-step denoising in diffusion models slows down these methods, and the generated video often lacks 3D constraints, resulting in structural inconsistencies and unrealistic motion. To address these limitations, we propose the VideoScene framework, which uses 3D-aware leap flow distillation to incorporate 3D priors for consistent 3D video generation while accelerating diffusion denoising in a single step for fast, high-quality generation.
\section{Method}
\begin{figure*}[!t]
    \centering
    \includegraphics[width=\linewidth]{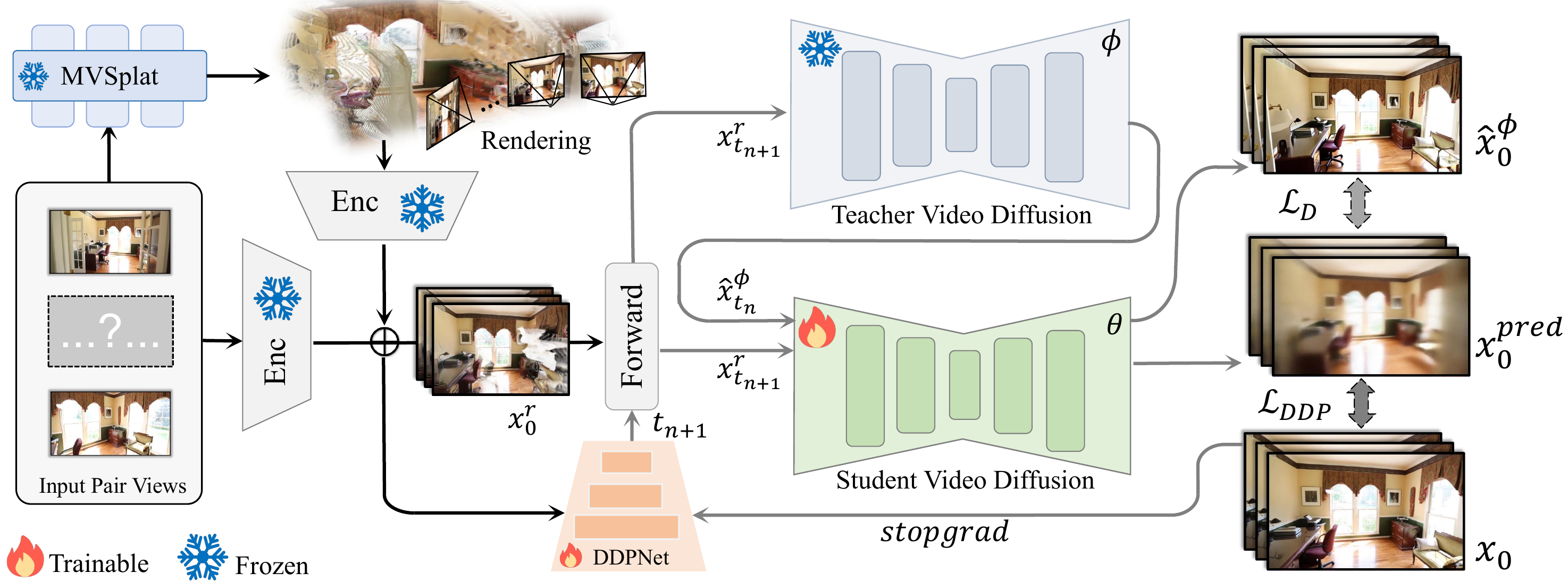}
    \caption{\textbf{Pipeline of VideoScene.} Given input pair views, we first generate a coarse 3D representation with a rapid feed-forward 3DGS model (\ie, MVSplat~\cite{chen2025mvsplat}), which enables accurate camera-trajectory-control rendering. The encoded rendering latent (``input'') and encoded input pairs latent (``condition'') are combined as input to the consistency model. Subsequently, a forward diffusion operation is performed to add noise to the video. Then, the noised $\bx_{n+1}^r$ is sent to both the student and teacher model to predict videos $\bx_0^{pred}$ of timestep $t_{n+1}$ and $\hat{\bx}_0^{\phi}$ of timestep $t_n$. Finally, the student model and DDPNet are updated independently through distillation loss and DDP loss.}
    % \vspace{-0.5cm}
    \label{fig:pipeline}
\end{figure*}

\subsection{Preliminaries}
\label{sec:preliminaries}
By interpreting the guided reverse diffusion process as solving an augmented probability flow ODE (PF-ODE)~\cite{song2020score-based, lu2022dpm-solver}, consistency model~\cite{song2023consistency_model} offers a generative approach that supports efficient one-step or few-step generation by directly mapping noise to the initial point of PF-ODE trajectory (\ie, the solution of the PF-ODE). This mapping is achieved through a consistency function, defined as $\mathbf{f}:(\bx_t,t) \longmapsto \bx_\epsilon$, where $\epsilon$ is a fixed small positive number, $t$ is the current denoising step, and $\bx_t$ represents the noisy input. One important self-consistency property that the consistency function should satisfy can be formulated as:
\begin{equation}
    \mathbf{f}(\bx_t,t)=\mathbf{f}(\bx_{t'},t'),\quad \forall t,t'\in[\epsilon,T],
    \label{eq:self_consistency}
\end{equation}
where $T$ donates the overall denoising step. This property ensures that the function outputs remain consistent across different denoising steps within the defined interval. The consistency function is parameterized as a deep neural network $\mathbf{f_\theta}$, with parameters $\mathbf{\theta}$ to be learned through the Consistency Distillation. The distillation goal is to minimize the output discrepancy between random adjacent data points, thereby ensuring self-consistency in the sense of probability. Formally, the consistency loss is defined as follows:
\begin{small}
\begin{equation}
    \mathcal{L}(\mathbf{\theta}, \mathbf{\theta}^-;\Phi)=\mathbb{E}_{\boldsymbol{x},t} 
    \left[d\left(\mathbf{f}_\mathbf{\theta}(\bx_{t_{n+1}},t_{n+1}),\mathbf{f}_{\mathbf{\theta}^{-}}(\hat{\bx}^\phi_{t_n},t_n)\right)\right],
    \label{eq:consistency_loss}
\end{equation}
\end{small}
\normalsize where $\Phi(\cdot)$ denotes the one-step ODE solver applied to PF-ODE, the model parameter $\mathbf{\theta^-}$ is obtained from the exponential moving average (EMA) of $\mathbf{\theta}$, and $d(\cdot,\cdot)$ is a chosen metric function for measuring the distance between two samples. Here, $\hat{\bx}^\phi_{t_n}$ is the estimation of $\bx_{t_n}$ from $\bx_{t_{n+1}}$:
\begin{equation}
    \hat{\bx}_{t_n}^\phi\leftarrow \bx_{t_{n+1}}+(t_n-t_{n+1})\Phi(\bx_{t_{n+1}},t_{n+1};\phi).
    \label{eq:one_step_solver}
\end{equation}
LCM~\cite{luo2023latent_consistency_model} conducts the above consistency optimization in the latent space and applies classifier-free guidance~\cite{ho2022classifier} in Eq.~\ref{eq:one_step_solver} to inject control signals, such as textual prompts. $\hat{\bx}_{t_n}^\phi$ is estimated by a teacher model with an ODE solver in consistency distillation. 
With a well-trained consistency model $\mathbf{f}_\mathbf{\theta}$, we can generate samples by sampling from the initial distribution $\bx_T \sim \mathcal{N}(\mathbf{0}, T^2 \boldsymbol{I})$ and then evaluating the consistency model for $\bx_\epsilon=\mathbf{f}_\mathbf{\theta}(\bx_T,T)$. Ideally, this involves only one forward pass through the consistency model and therefore generates samples in a single step. 

\subsection{Challenges in Video to 3D}
Video diffusion models have shown impressive results and great potential, especially in 3D reconstruction. However, when generating videos in a static scene, undesirable variations often arise~\cite{zeng2024dawn}—such as human motions, object interactions, and environmental changes—that compromise the 3D consistency required for reliable reconstruction. Our goal is to ensure 3D consistency in generated videos with high efficiency, producing an effective tool to bridge the gap from video to 3D. In other words, we aim for outputs that adhere to a 3D-consistent data distribution while excluding disruptive variations. A straightforward approach to address this is fine-tuning the diffusion model on a dedicated 3D dataset, forcing the model to align with a 3D consistent distribution. This method has proven effective in various style transfer tasks, which similarly aim to preserve a specific data distribution within a larger, more diverse set. However, this method presents two main challenges. First, fine-tuning does not enhance diffusion efficiency: achieving high-quality video output still requires up to 50 denoising steps, which is computationally costly and time-intensive. Second, the generation process lacks controllability. The model, though fine-tuned, follows a standard diffusion denoising procedure, starting from a random distribution and gradually reaching the target data distribution via step-by-step denoising. This stochastic process makes it difficult to enforce a consistent camera trajectory and 3D coherence, even when given image conditions.

To address these challenges, we introduce VideoScene, a novel distillation technique for one-step, 3D-consistent video generation to bridge the gap from video to 3D. Specifically, our approach incorporates a 3D-aware leap flow distillation strategy (Sec.~\ref{sec:distillation}) and a dynamic denoising policy network (Sec.~\ref{sec:policy_network}) during both training and inference to maximize 3D priors, enhancing both efficiency and controllability of video generation.

\subsection{3D-Aware Leap Flow Distillation}
\label{sec:distillation}
% 1. Initial sample in 3D domain, also end in 3D domain to ensure consistency in trajectory
% 2. Why not in GT distribution?
In the consistency distillation training~\cite{luo2023latent_consistency_model}, a conventional noise scheduler samples an initial ground truth $\bx_0$ from the data distribution and applies noise to generate $\bx_t$ at a random timestep $t$ using forward diffusion as follows:
\begin{equation}\label{eq:forward}
\bx_t = q(\bx_0, \boldsymbol{\epsilon}, t) = \alpha_t \bx_0 + \sigma_t \boldsymbol{\epsilon},\quad \forall t\in [0,T],
\end{equation}
where Gaussian noise $\boldsymbol{\epsilon} \sim \mathcal{N}(0,\textbf{I})$, $\alpha_t$ and $\sigma_t$ define the signal-to-noise ratio (SNR) of the stochastic interpolant $\bx_t$. Standard SNR schedules ensure that $\bx_T$ retains some low-frequency information from $\bx_0$, causing a mismatch with inference, which starts from a fully noisy $\bx_T$. This mismatch leads to degradation during inference, especially with fewer denoising steps. While some studies~\cite{lin2024common, kohler2024imagine_flash} suggest adjusting the noise schedule or applying backward distillation to maintain zero terminal SNR, we find these solutions insufficient for efficient performance.

Observing that the initial denoising steps (where $t$ is near $T$) are particularly challenging due to limited prior information, we propose a 3D-aware leap flow distillation strategy that aligns inference with training at an intermediate timestep $t,t\in [0, T']$, where $T'<T$. Specifically, during distillation training, we first employ a fast, feedforward sparse-view 3DGS~\cite{kerbl2023_3dgs} model, MVSplat~\cite{chen2025mvsplat}, to generate a coarse but 3D-consistent scene by matching and fusing view information with the cross-view transformer and cost volume. Then we render continuous frames from the coarse scene as follows:
\begin{equation}
    \{I_\mathrm{Render}\}_{\tau=1}^\mathcal{T}=g(S(I_\mathrm{Input}^i,c^i), o(c^i)),\quad i=\{0,1\}
    \label{eq:render}
\end{equation}
where \(I_\mathrm{Input}\) represents the pairwise input image, $c^i$ is the corresponding camera pose, $S(\cdot,\cdot)$ is the sparse-view reconstruction model, $o(\cdot)$ is the interpolation camera trajectory, $g(\cdot,\cdot)$ is a renderer given 3D representation and queried camera trajectory, and \(\{I_\mathrm{Render}\}_{\tau=1}^\mathcal{T}\) is the rendered video. Although the video shows visual artifacts and blurred regions, it contains the scene’s 3D geometric structure information as rendered from a 3D representation. From it, we encode the rendered video $\{I_\mathrm{Render}\}_{\tau=1}^\mathcal{T}$ into latent space and sample a $\bx_0^r$, adding noise at a randomly selected timestep $t$ according to Eq.~\ref{eq:forward}, and training gradients are calculated as follows:
\begin{small}
\begin{equation}
    \mathcal{L}_{D}(\mathbf{\theta}, \mathbf{\theta}^-;\Phi)=\mathbb{E}_{\boldsymbol{x},t} 
    \left[d\left(\mathbf{f}_\mathbf{\theta}(\bx_{t_{n+1}}^r,t_{n+1}),\mathbf{f}_{\mathbf{\theta}^{-}}(\hat{\bx}^\phi_{t_n}(\bx_{t_{n+1}}^r),t_n)\right)\right],
    \label{eq:distillation_loss}
\end{equation}
\end{small}
where $\hat{\bx}^\phi_{t_n}(\bx_{t_{n+1}}^r)$ is estimated by the teacher model with input $\bx_{t_{n+1}}^r$ following Eq.~\ref{eq:one_step_solver} and fed into the EMA student model. To make it easier to understand, we refer to both the student and EMA student as one "student video diffusion" in Fig.~\ref{fig:pipeline}. During inference, we also start from $\bx_0^r$ and add noise at a selected timestep $t$, and this selection follows a policy network rather than a random approach. This will be discussed in Sec~\ref{sec:policy_network}.

In summary, our 3D-aware leap flow distillation mitigates discrepancies between training and inference by avoiding reliance on ground-truth signals and bypassing inefficient early denoising steps near $T$. This process accelerates overall denoising by effectively simulating training during inference, enabling the student model to leverage rich prior knowledge rather than starting from scratch.

\subsection{Dynamic Denoising Policy Network}
\label{sec:policy_network}
During inference, we begin from the initial rendered video latents $\bx_0^r$, and progressively add noise at a chosen leap timestep $t$. The choice of $t$ is context-dependent: when the input video is of high quality, adding a small amount of noise suffices to refine fine details while preserving overall structure. However, when input videos contain artifacts such as structural distortions, blurring, or lighting inconsistencies, a larger noise addition is necessary as a minimal noise step could introduce an unsuitable prior, resulting in degraded structure. Conversely, excessive noise risks overwhelming the model’s prior, yielding outputs close to pure noise and losing critical 3D information. Therefore, selecting an appropriate denoising timestep is essential for optimal inference performance.

To better decide the noise level, we introduce a policy network based on a contextual bandit algorithm~\cite{chu2011contextual, pan2019policy, bouneffouf2020survey, shi2023deep}. This network acts as the agent and learns to dynamically select the best timestep $t$ for denoising. We model this $t$ selection as an independent decision process: given an environment state (\ie, input video latent $\bx_0^r$), the agent (\ie, DDPNet with policy distribution $\pi_\psi(t|\bx_0^r;\psi)$) decides an action (\ie, a noise step $t \in [0, T’]$) and receive a reward (\ie, loss $\mathcal{L}_{MSE}$). At each distillation training round, the decision of $t$ is made randomly and applied to an input latent $\bx_0^r$. The sample data estimate $\bx_0^{pred}(t)$ can be computed as:
\begin{equation}
    \bx_0^{pred}(t)=\frac{\bx_t^r-\sigma_t \mathbf{\epsilon}_\mathbf{\theta}(\bx_t^r, t)}{\alpha_t},
\end{equation}
where $\mathbf{\epsilon}_\mathbf{\theta}(\cdot,\cdot)$ is the noise predictor of student diffusion model. After denoising, the predicted video output $\bx_0^{pred}$ is compared to the ground truth sequence $\bx_0$ using MSE loss:
\begin{equation}
    \mathcal{L}_{MSE}=stopgrad(\Vert \bx_0^{pred}(t)-\bx_0 \Vert_2^2).
    \label{eq:mse_loss}
\end{equation}
The grad is stopped here as the MSE loss is only used to update the policy network. We define the immediate reward $r(\bx_0^r, t)=-\mathcal{L}_{MSE}$, providing a direct optimization target for the policy network. This reward signal encourages timestep selections that yield minimal reconstruction error. We achieve this through policy gradient optimization~\cite{sutton1999policy, pan2019policy}, which adjusts $\psi$ to increase the likelihood of timestep selections associated with higher rewards by minimizing $\mathcal{L}_{DDP}(\psi)$:
\begin{equation}
    \mathcal{L}_{DDP}(\psi)=\mathbb{E}_{t \sim \pi_\psi(t|\bx_0^r;\psi)}[r(\bx_0^r, t)],
    \label{eq:policy_loss}
\end{equation}
It should be noted that during the training of the policy network, data from the distillation training is exclusively utilized by the policy network. To ensure stability in the distillation process, the policy network does not pass any gradients to the student model.

\begin{figure*}[!t]
    \centering
    \includegraphics[width=\linewidth]{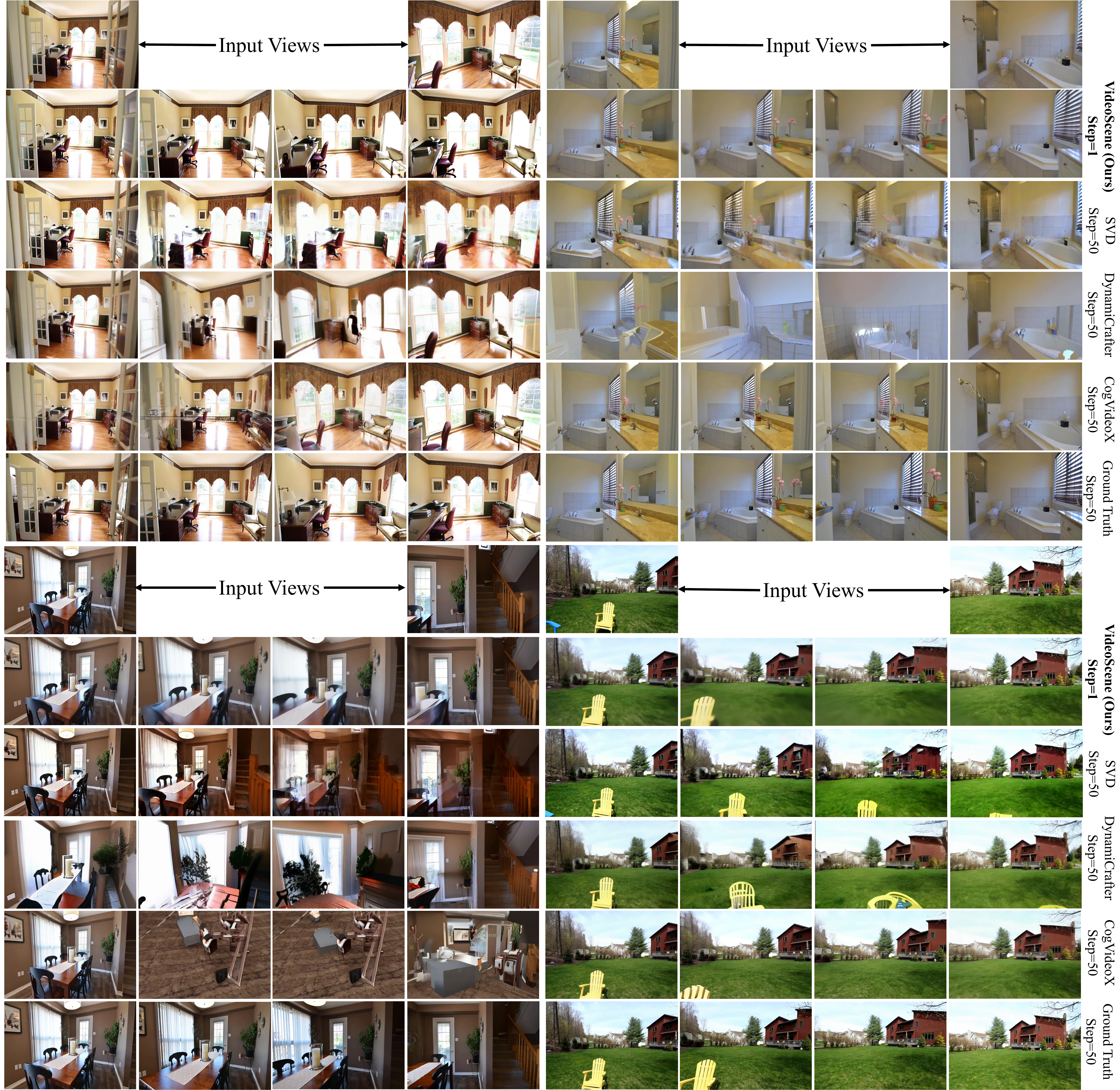}
    \caption{\textbf{Qualitative comparison.} We can observe that baseline models suffer from issues such as blurriness, frame skipping, excessive motion, and shifts in the relative positioning of objects, while our VideoScene achieves higher output quality and improved 3D coherence.}
    \label{fig:comparison}
    \vspace{-3mm}
\end{figure*}
\section{Experiments}
\subsection{Experiment Setup}
\textbf{Implementation Details.} To implement the proposed VideoScene, we choose MVSplat~\cite{chen2025mvsplat} as feed-forwad 3DGS model and a pretrained CogVideoX-5B-I2V~\cite{yang2024cogvideox} (@ 720 $\times$ 480 resolution) as the video diffusion backbone. The pretrained CogVideoX costs more than 2 minutes with default 50 DDIM~\cite{song2020ddim} steps. We first finetune the attention layers of the video model with 900 steps on the learning rate $1 \times 10^{-4}$ for warm-up. Then we leverage the pre-trained 3D model with fixed parameters and conduct distillation training for 20k iterations. Our video diffusion was trained on 3D scene datasets RealEstate10K~\cite{zhou2018real10k} by sampling 49 frames with a batch size of 2. During training, only the attention layers within the transformer blocks of the video model are updated. The training is conducted on 8 NVIDIA A100 (80G) GPUs in two days, using the AdamW~\cite{loshchilov2017adamw} optimizer with a learning rate of $3\times 10^{-5}$. The dynamic denoising policy network (DDPNet) follows a CNN architecture, with 4 layers of 2D convolution along with corresponding normalization and activation layers. Since the policy network has significantly fewer parameters than the video diffusion model, it participates in full training only during the first 4,000 steps to prevent overfitting. Notice that the inference time of our VideoScene only costs within 3s (renderings from 3DGS feed-forward model~\cite{chen2025mvsplat} cost $\sim$ 0.5s and the distilled video generation with DDIM~\cite{song2020ddim} sampler in one step costs $\sim$ 2.5s).

\begin{table*}[t]
\scriptsize
    \centering
    \caption{\textbf{Quantitative Comparison.} We compare 1-, 4-, and 50-step versions of models. Not only does VideoScene outperform other methods, but its 1-step results also remain closely comparable to the 50-step results, while other methods show a significant decrease.}
    \vspace{-0.1cm}
    \resizebox{1\linewidth}{!}{
    \begin{tabular}{lcccccc}
    \toprule
    {Method} & {\#Steps} & FVD $\downarrow$ & {Aesthetic Quality} $\uparrow$ & Subject Consistency $\uparrow$ & {Background Consistency} $\uparrow$ \\
    \midrule
    \multirow{3}{*}{Stable Video Diffusion~\cite{blattmann2023svd}} 
        & 50 & 424.68 & 0.4906 & 0.9305 & 0.9287 \\ 
        & 4 & 541.89 & 0.4040 & 0.8728 & 0.8952 \\
        & 1 & 1220.80 & 0.2981 & 0.7934 & 0.8817 \\ 
    \midrule
    \multirow{3}{*}{DynamiCrafter~\cite{xing2025dynamicrafter}} 
        & 50 & 458.27 & 0.5336 & 0.8898 & 0.9349 \\ 
        & 4 & 512.50 & 0.4899 & 0.8661 & 0.9098 \\
        & 1 & 846.85 & 0.3737 & 0.7474 & 0.8627 \\
    \midrule
    \multirow{3}{*}{CogVideoX-5B~\cite{yang2024cogvideox}} 
        & 50 & 521.04 & 0.5368 & 0.9179 & 0.9460 \\ 
        & 4 & 662.13 & 0.4486 & 0.8489 & 0.9168 \\
        & 1 & 753.02 & 0.3987 & 0.7842 & 0.8976 \\
    \midrule
    \multirow{3}{*}{\textbf{VideoScene (Ours)}} & 50 & \textbf{98.67} & \textbf{0.5570} & \textbf{0.9320}  & \underline{0.9407}   \\
    & 4 & 175.84 & \underline{0.5357}  & \underline{0.9269} &  \underline{0.9431} \\
    & 1 & \underline{103.42} & \underline{0.5416}  & \underline{0.9259} & \textbf{0.9461}  \\

    \bottomrule
    \end{tabular}}
    \vspace{-0.2cm}
    \label{tab:quantitative-comparison}
\end{table*}

\noindent\textbf{Datasets.} We assess VideoScene on the large-scale 3D scene dataset RealEstate10K~\cite{zhou2018real10k}. RealEstate10K is a dataset downloaded from YouTube, which is split into 67,477 training scenes and 7,289 test scenes. To better verify the effectiveness of VideoScene, we have established a challenging benchmark, testing on 120 benchmark scenes with large angle variance. The dataset also provides estimated camera intrinsic and extrinsic parameters for each frame. We first fine-tune a video interpolation model with the first and end frame guidance. For each scene video, we sample 49 contiguous frames with equal intervals and serve the first and last frames as the input for our video diffusion model. To further evaluate our cross-dataset generalization ability, we also evaluate on the video dataset ACID~\cite{liu2021acid}, which contains natural landscape scenes with camera poses.

\noindent\textbf{Baselines and Metrics.} As the fundamental unit of multi-view geometry is two-view geometry, our goal is to use video diffusion to generate new, 3D-consistent viewpoints between the given two input views. To evaluate our approach, we compare it against several state-of-the-art open-source video frame interpolation models: Stable Video Diffusion~\cite{blattmann2023svd}, DynamiCrafter~\cite{xing2025dynamicrafter}, and CogVideoX~\cite{yang2024cogvideox}. Stable Video Diffusion and DynamiCrafter employ a UNet-based diffusion architecture, while CogVideoX utilizes a diffusion transformer architecture. For quantitative evaluation, we assess the quality and temporal coherence of synthesized videos by reporting the Fréchet Video Distance (FVD)~\cite{unterthiner2018fvd}. Following the previous benchmark VBench~\cite{huang2023vbench}, we also evaluate the Aesthetic Score, Subject Consistency, and Background Consistency of generated videos as our metrics. See supplementary for more details.

\subsection{Comparison with Baselines}
We compare our VideoScene framework against three baseline models across different DDIM steps (1, 4, and 50) in Tab.~\ref{tab:quantitative-comparison}. Visual comparisons of our one-step results versus the 50-step results of baselines are shown in Fig.~\ref{fig:comparison}. Our VideoScene outperforms state-of-the-art models, even with one-step denoising, achieving superior visual quality and spatial consistency with the ground truth. More comparison results are provided in the supplement material.

\begin{table*}[!t]
    \centering
    \caption{\textbf{Quantitative results in cross-dataset generalization}. Models trained on the source dataset RealEstate10K~\cite{zhou2018real10k} are tested on unseen scenes from target datasets ACID~\cite{liu2021acid}, without any fine-tuning.}
    % \vspace{-1em}
    \resizebox{1\linewidth}{!}{
    \begin{tabular}{lcccccc}
    \toprule
    Method & 3D Training data & \#Steps & FVD$\downarrow$ & {Aesthetic Quality} $\uparrow$ & Subject Consistency $\uparrow$ & {Background Consistency} $\uparrow$ \\
    \midrule 
    \multirow{2}{*}{DynamiCrafter~\citep{xing2025dynamicrafter}} & \multirow{2}{*}{/} & 50 &  242.76&  0.5191&  0.9464& 0.9527\\ 
    && 1 &  453.24&  0.4126&  0.8136& 0.8841\\ 
    \midrule
    \multirow{2}{*}{CogVideoX-5B~\citep{yang2024cogvideox}} & \multirow{2}{*}{/} & 50 &  867.31&  0.5212&  0.9628& \textbf{0.9720} \\ 
    & & 1 &  537.48&  0.4614&  0.8452& 0.9349\\
    \midrule 
    \multirow{2}{*}{DynamiCrafter~\citep{xing2025dynamicrafter}} & \multirow{2}{*}{RealEstate10K~\citep{zhou2018real10k}} & 50 & 96.10 &  0.5096&  0.9524&   0.9545\\ 
    && 1 & 595.87&  0.4013&  0.8202&   0.8917\\ 
    \midrule
    \multirow{2}{*}{CogVideoX-5B~\citep{yang2024cogvideox}} & \multirow{2}{*}{RealEstate10K~\citep{zhou2018real10k}} & 50 &  114.04 &  0.5491&  \textbf{0.9637} &   0.9593\\ 
    & & 1 &  464.87&  0.4492&  0.8406&   0.9214\\ 
    \midrule
    \multirow{2}{*}{\textbf{VideoScene (Ours)}} & \multirow{2}{*}{RealEstate10K~\citep{zhou2018real10k}} & 50 & \textbf{73.93} & \textbf{0.5602} & \underline{0.9598} & \underline{0.9573} \\ 
    && 1 & \underline{121.93} & \underline{0.5274} & \underline{0.9395} & \underline{0.9494} \\ 
    \bottomrule
    \end{tabular}}
    \vspace{-0.2cm}
    \label{tab:generalization}
\end{table*}

\begin{figure}[ht]
    \centering
    \includegraphics[width=\linewidth]{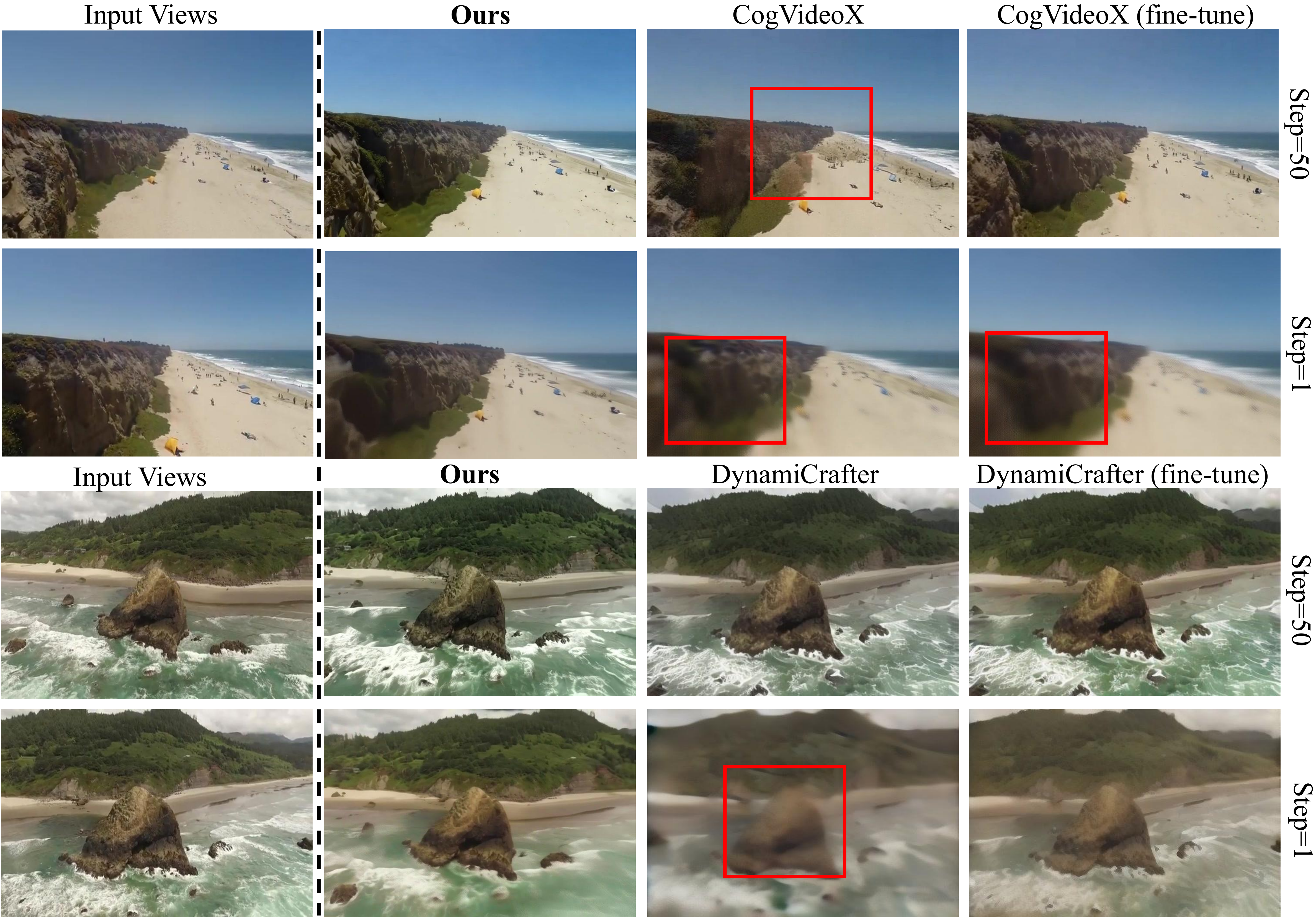}
    \vspace{-0.3cm}
    \caption{\textbf{Qualitative results in cross-dataset generalization.} Models trained on the source dataset RealEstate10K are tested on ACID dataset. Fine-tuned models improve in 3D consistency but still fail with one-step.}
    \vspace{-0.3cm}
    \label{fig:cross-data}
\end{figure}

\noindent\textbf{Cross-dataset generalization.} Leveraging the strong generative capabilities of the video diffusion model with 3D structure prior, VideoScene demonstrates a natural advantage in generalizing to novel, out-of-distribution scenes shown in Fig.~\ref{fig:teaser}. To validate this generalizability, we conduct a cross-dataset evaluation. For a fair comparison, we train baseline models (DynamiCrafter~\cite{xing2025dynamicrafter} and CogVideoX~\cite{yang2024cogvideox} included) on the same 3D dataset~\cite{zhou2018real10k} and directly test them on the ACID~\cite{liu2021acid} dataset. As shown in Tab.~\ref{tab:generalization} and Fig.~\ref{fig:cross-data}, while baselines improve 3D consistency with 50-step inference after fine-tuning on 3D data, they fail to achieve clarity with one-step inference. Remarkably, our method not only achieves comparable performance to the baselines' 50-step results but also surpasses them in one-step inference, highlighting its strong generalizability.

\begin{figure}[ht]
    \centering
    % \vspace{-0.3cm}
    \includegraphics[width=\linewidth]{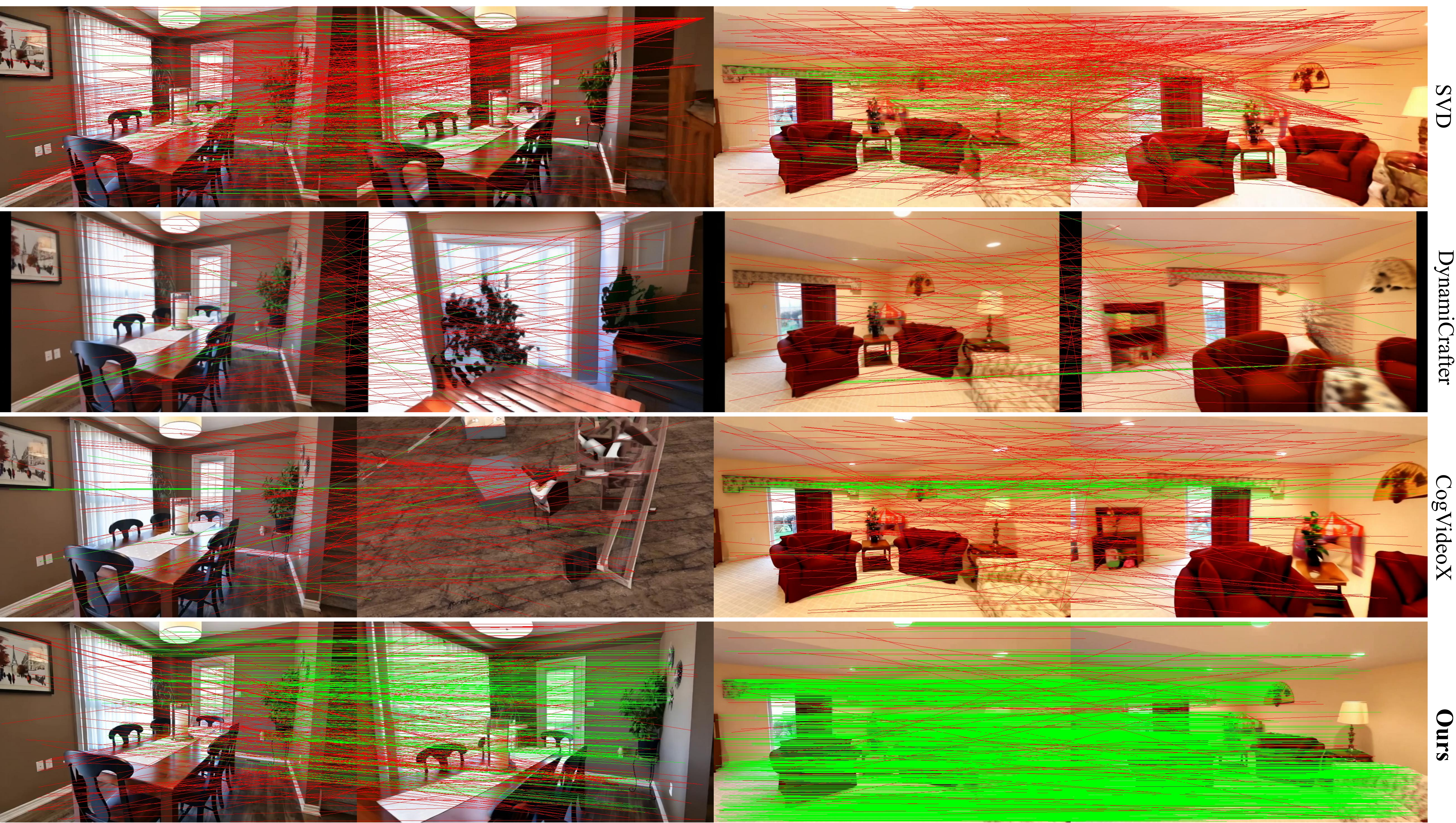}
    % \vspace{-0.6cm}
    \caption{\textbf{Matching results comparison.} Green represents high-quality matching results, while red represents discarded matching results. More green high-quality matches indicate a higher level of geometric consistency between the two views.}
    \label{fig:match}
\end{figure}

\noindent\textbf{Structure matching comparison.} To further assess geometric consistency, we evaluate the camera geometry alignment between frames in the generated videos following~\cite{li2024sora_geo}. Specifically, we extract two frames at regular intervals from each generated video, creating pairs of two-view images. For each pair, we apply a feature-matching algorithm~\cite{ng2003sift} to find corresponding points and use RANSAC~\cite{fischler1981ransac} with the fundamental matrix (epipolar constraint) to filter out incorrect matches. Fig.~\ref{fig:match} shows that VideoScene achieves the highest number of correctly matched points, confirming superior geometry consistency.

\begin{table}[!h]
    \centering
    \caption{\textbf{Quantitative results of ablation study}. We report the quantitative metrics ablations in RealEstate10K.}
    \vspace{-0.1cm}
    \resizebox{1\linewidth}{!}{
    \begin{tabular}{lcccc}
        \toprule
        Setup  & FVD$\downarrow$ & {Aesthetic Quality} $\uparrow$ & Subject Consistency $\uparrow$ & {Background Consistency} $\uparrow$ \\
        \midrule
        Base rendered video &  171.38&  0.4769&  0.8794& 0.9240\\
        w/o 3D-aware leap flow &  543.53&  0.4092&  0.7842& 0.9160\\
        w/o DDPNet &  106.28 &  0.4897 &  0.8850 & 0.9205\\
        \textbf{VideoScene (full model)} & \textbf{97.53} & \textbf{0.5306}  & \textbf{0.9139} & \textbf{0.9440}\\
        \bottomrule
    \end{tabular}}
    \vspace{-0.2cm}
    \label{tab:ablation}
\end{table}
\subsection{Ablation Study and Analysis}
\begin{figure}
    \centering
    % \vspace{-0.4cm}
    \includegraphics[width=\linewidth]{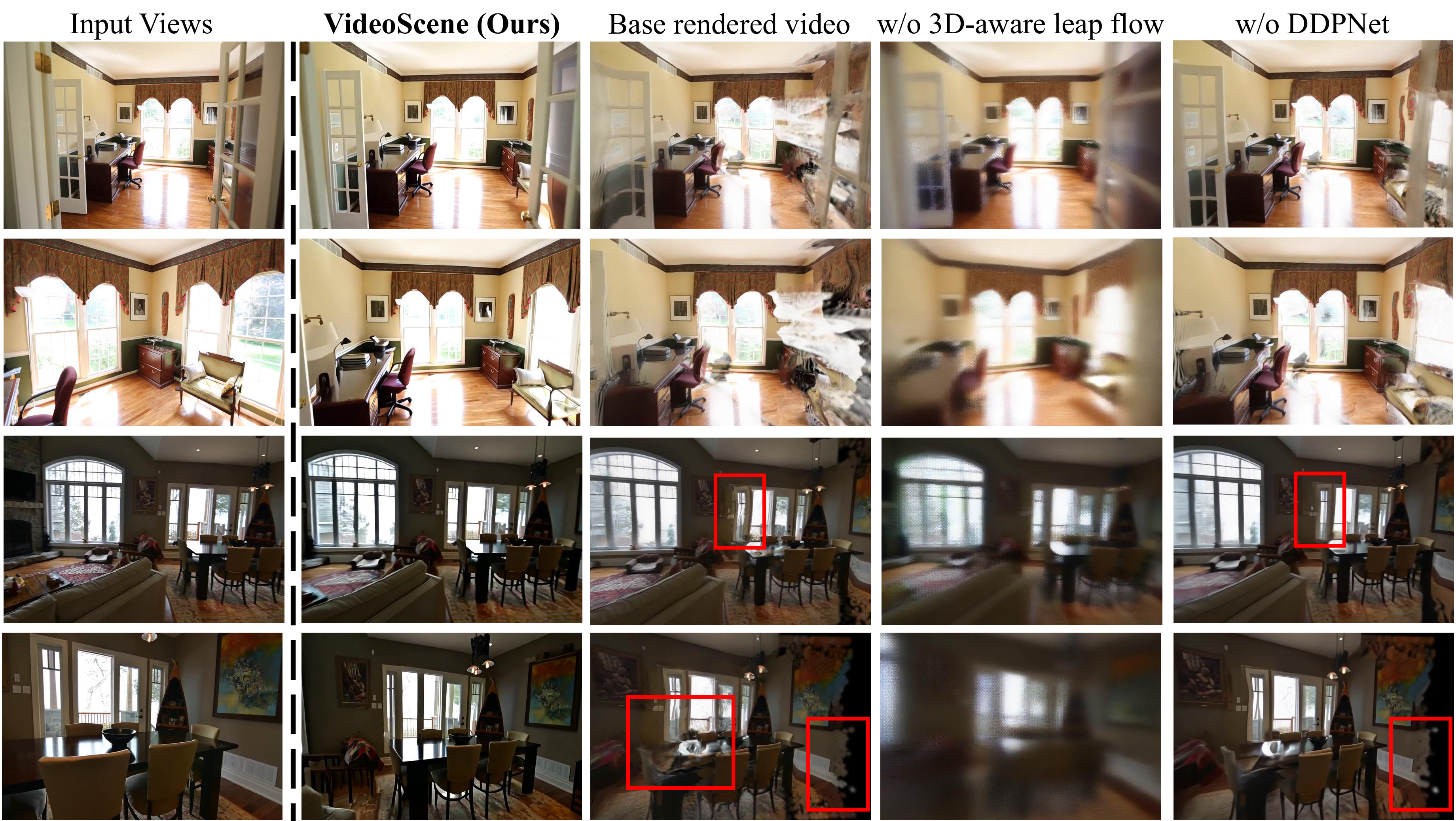}
    % \vspace{-0.6cm}
    \caption{\textbf{Visual results of ablation study.} We ablate the design choices of 3D-aware leap flow distillation and dynamic denoising policy network (DDPNet).}
    % \vspace{-0.4cm}
    \label{fig:ablation}
\end{figure}

We perform ablation studies to analyze the design choices within the VideoScene framework (see Tab.~\ref{tab:ablation} and Fig.~\ref{fig:ablation}). The base rendered video represents the video generated directly from the 3D model. A naive combination of a 3D-fine-tuned video diffusion model with standard distillation acceleration methods~\cite{wang2023videolcm} is referred to as “w/o 3D-aware leap flow.” Additionally, we perform ablation on DDPNet. Results indicate that while the rendered video from the 3D model has suboptimal quality, it provides coarse consistent information. Without 3D-aware leap flow distillation, generated frames suffer from inconsistencies, leading to blur and artifacts. 
The inclusion of DDPNet further enhances fine-grained details and corrects spatial distortions, demonstrating its effectiveness in optimal denoising decisions.

% \section{Conclusion}
% This paper introduces VideoScene, a novel fast video generation framework that distills the video diffusion model to generate 3D scenes in one step. The key to our method is that we constrain the optimization with 3D prior and propose a 3D-aware leap flow distillation strategy to leap over time-consuming redundant information. Moreover, we design a dynamic denoising policy network to adaptively determine the optimal leap timestep during inference. Extensive experiments demonstrate the superiority of our proposed VideoScene in terms of efficiency and consistency in 3D structure, highlighting its potential as an efficient and effective tool to bridge the gap from video to 3D.
\section{Conclusion}
In this paper, we introduce VideoScene, a novel fast video generation framework that distills the video diffusion model to generate 3D scenes in one step. Specifically, we constrain the optimization with 3D prior and propose a 3D-aware leap flow distillation strategy to leap over time-consuming redundant information. Moreover, we design a dynamic denoising policy network to adaptively determine the optimal leap timestep during inference. Extensive experiments demonstrate the superiority of our proposed VideoScene in terms of efficiency and consistency in 3D structure, highlighting its potential as an efficient and effective tool to bridge the gap from video to 3D.

% In this paper, we introduce VideoScene, a novel fast video generation framework that distills the video diffusion model to generate 3D scenes in one step. The key to our method is that we constrain the optimization with 3D prior and propose a 3D-aware leap flow distillation strategy to leap over time-consuming redundant information. Moreover, we design a dynamic denoising policy network to adaptively determine the optimal leap timestep during inference. Extensive experiments demonstrate the superiority of our proposed VideoScene in terms of efficiency and consistency in 3D structure, highlighting its potential as an efficient and effective tool to bridge the gap from video generation to 3D reconstruction.
% \newline
% \newline
% \noindent \textbf{Acknowledgements.} 
% The work was supported in part by the National Natural Science Foundation of China under Grant 62206147.

{
    \small
    \bibliographystyle{ieeenat_fullname}
    \bibliography{main}
}

\clearpage
\setcounter{page}{1}
\maketitlesupplementary

\section{More Discussion of Preliminaries} 
In this section, we provide more preliminaries about the diffusion model, consistency model~\cite{song2023consistency_model}, and contextual bandit~\cite{thompson1933bandit}.

\subsection{Diffusion Model}
Diffusion models~\cite{ho2020ddpm,song2020ddim,song2020score-based,rombach2022ldm} generate data by progressively introducing Gaussian noise to the original data and subsequently sampling from the noised data through several denoising steps.
Let $p_{\mathrm{data}}(x)$ denote the data distribution,  The forward process is described by a stochastic differential equation (SDE)~\cite{song2020score-based} given by
\begin{equation}
    \mathrm{d}\mathbf{x}_t = \mathbf{\mu}(\mathbf{x}_t, t)\mathrm{d}t
    + \sigma(t)\mathrm{d}\mathbf{w}  
\end{equation}
where $t \in [0, T]$, $T > 0$ denotes a fixed time horizon, $\mathbf{\mu}(\cdot, \cdot)$ and $\sigma(\cdot)$ represent the drift and diffusion coefficients,respectively, and $\{\mathbf{w_t}\}_{t \in [0, T]}$ is the standard Brownian motion. 
An important property of this SDE is the existence of an associated ordinary differential equation (ODE), known as the Probability Flow (PF) ODE~\cite{song2020score-based}, which deterministically describes the distribution's evolution
\begin{equation}
    \mathrm{d}\mathbf{x}_t = \left[\mu(\mathbf{x}_t, t) - \frac{1}{2} \sigma^2(t) \nabla_{\mathbf{x}_t} \log p_t(\mathbf{x}_t)\right] \mathrm{d}t
    \label{eq: forward}
\end{equation}
where $\nabla_{\mathbf{x}_t} \log p_t(\mathbf{x}_t)$ is the  \textit{score function} of the intermediate distribution $p_t(\mathbf{x}_t)$. 
For practical purposes~\cite{karras2022elucidating}, a simplified setting is often adopted, where $\mathbf{\mu}(\mathbf{x}_t, t) = \mathbf{0}$
and $\sigma(t) = \sqrt{2t}$. This yields the intermediate distributions $p_t(\mathbf{x}) = p_{\mathrm{data}}(x) \otimes \mathcal{N}(\mathbf{0}, t^2\mathbf{I})$, where $\otimes$ convolution operation. Let $\pi(\mathbf{x}) = \mathcal{N}(0, T^2\mathbf{I})$ and after a sufficient noise adding process, the final distribution $p_T(\mathbf{x})$ will be closed to $\pi(\mathbf{x})$. Sampling involves solving the empirical PF ODE:
\begin{equation}
    \frac{d\mathbf{x}_t}{dt} = -t \nabla_{\mathbf{x}_t} \log p_t(\mathbf{x}_t)
\end{equation}
starting from a sample $\mathbf{x}_T \sim \mathcal{N}(\mathbf{0}, T^2\mathbf{I})$ and running the ODE backward procedure with Numerical ODE solver like  Euler~\cite{song2020denoising} and Heun~\cite{karras2022elucidating} solver, we can obtain a solution trajectory $\{\hat{\mathbf{x}}_t\}_{t \in \left[0, T\right]}$ and thus get a approximate sample $\hat{\mathbf{x}}_0$ from the data distribution $p_{\mathrm{data}}(\mathbf{x})$. The backward process is typically stopped at $t = \epsilon$ to avoid numerical instability, where $\epsilon$ is a small positive number, and $\hat{\mathbf{x}}_{\epsilon}$  is treated as the final approximate result.

\subsection{Consistency Model}
Consistency model~\cite{song2023consistency_model} is a novel class of models that supports both one-step and iterative generation, providing a trade-off between sample quality and computational efficiency. The consistency model can be trained either by distilling knowledge from a pre-trained diffusion model or independently, without relying on pre-trained models.
Formally, given a solution trajectory $\{\hat{\mathbf{x}}_t\}_{t \in \left[0, T\right]}$ sampled from Eq.~\ref{eq: forward}, we define the \textit{consistency function} as $\mathbf{f}: (\mathbf{x}_t, t) \mapsto \mathbf{x}_{\epsilon}$. a consistency function exhibits a self-consistency property, meaning that its outputs remain consistent for any pair of
$(\mathbf{x}_t, t)$ points that lie along the same PF ODE trajectory. The goal of a consistency model is to approximate this consistency function $\mathbf{f}$ with $\mathbf{f}_{\mathbf{\theta}}$. Given any consistency function \( \mathbf{f}(\cdot, \cdot) \), it must satisfy \( \mathbf{f}(x_\epsilon, \epsilon) = x_\epsilon \), implying that \( \mathbf{f}(\cdot, \epsilon) \) acts as the identity function. This requirement is referred to as the \textit{boundary condition}. It is imperative for all consistency models to adhere to this condition, as it is pivotal to the proper training of such models. There are several simple way to implement the \textit{boundary condition}, for example we can parameterize $\mathbf{f}_{\theta}$ as 
\begin{equation}
\mathbf{f}_\mathbf{\theta}(\mathbf{x}, t) = c_{\text{skip}}(t) \mathbf{x} + c_{\text{out}}(t) F_\mathbf{\theta}(\mathbf{x}, t)
\end{equation}
where \( c_{\text{skip}}(t) \) and \( c_{\text{out}}(t) \) are differentiable functions such that \( c_{\text{skip}}(\epsilon) = 1 \) and \( c_{\text{out}}(\epsilon) = 0 \). This parameterization ensures that the consistency model is differentiable at \( t = \epsilon \), provided that \( F_\mathbf{\theta}(x, t) \), \( c_{\text{skip}}(t) \), and \( c_{\text{out}}(t) \) are all differentiable, which is crucial for training continuous-time consistency models. 
Once a consistency model \( \mathbf{f}_\mathbf{\theta}(\cdot, \cdot) \) is well-trained, samples can be generated by first sampling from the initial distribution \( \hat{x}_T \sim \mathcal{N}(0, T^2 \mathbf{I}) \), and then evaluating the consistency model for \( \hat{x}_\epsilon = \mathbf{f}_\mathbf{\theta}(\hat{x}_T, T) \). This generates a sample in a single forward pass through the consistency model. Additionally, the consistency model can be evaluated multiple times by alternating between denoising and noise injection steps to improve sample quality, thus offering a flexible trade-off between computational cost and sample quality. This multi-step procedure also holds significant potential for zero-shot data editing applications.

\begin{algorithm*}[!t]
\begin{algorithmic}[1]
    \item \textbf{Input:} 3D dataset $\mathcal{D}$, initial model parameter $\theta$, learning rate $\eta$, one-step ODE solver $\Phi(\cdot)$, distance metric $d(\cdot,\cdot)$, EMA rate $\mu$, {noise schedule $\alpha_t,\sigma_t$, timestep interval $k$, diffusion optimization timesteps $T'$, and encoder $E(\cdot)$}
    \item \textbf{Repeat}
        \item \qquad Sample $\boldsymbol{\epsilon} \sim \mathcal{N}(0, I)$ and $t_{n+1} \in [0,T']$
        \item \qquad Sample  $(I_{Input}^i,c^i) \sim \mathcal{D}, \quad i=\{0,1\}$
        \item \qquad  $t_n \leftarrow t_{n+1} - k$
        \item \qquad Render images $\{I_\mathrm{Render}\}_{\tau=1}^\mathcal{T}=g(S(I_\mathrm{Input}^0,c^i), o(c^i)), \quad i=\{0,1\}$
        \item \qquad $\bx_0^r \leftarrow E(\{I_\mathrm{Render}\}_{\tau=1})$
        \item \qquad $\bx_{t_{n+1}}^r \leftarrow \alpha_{t_{n+1}} \bx_0^r + \sigma_{t_{n+1}} \boldsymbol{\epsilon}$
        \item \qquad $\hat{\bx}_{t_n}^\phi\leftarrow \bx_{t_{n+1}}^r+(t_n-t_{n+1})\Phi(\bx^r_{t_{n+1}},t_{n+1};\phi)$
        \item \qquad $\mathcal{L}_{D}(\mathbf{\theta}, \mathbf{\theta}^-;\Phi)\leftarrow d(\mathbf{f}_\mathbf{\theta}(\bx_{t_{n+1}}^r,t_{n+1}),\mathbf{f}_{\mathbf{\theta}^{-}}(\hat{\bx}^\phi_{t_n},t_n))$
        \item \qquad $\theta \leftarrow \theta - \eta \cdot \nabla_\theta\mathcal{L}_{D}(\theta,\theta^-;\Phi)$
        \item \qquad $\theta^-\leftarrow \text{stopgrad}(\mu\theta^-+(1-\mu)\theta)$
    \item \textbf{Until} convergence
\end{algorithmic}
\caption{3D-Aware Leap Flow Distillation}
\label{tab:algorithm}
\end{algorithm*}

\subsection{Contextual Bandit Algorithm}

The Multi-Armed Bandit (MAB) problem, originally introduced by~\cite{thompson1933bandit}, is a fundamental model in sequential decision-making under uncertainty. It is named after the analogy of a gambler trying to maximize rewards from multiple slot machines (or "arms"), each with an unknown probability distribution of payouts. At each step, the agent must decide which arm to pull, aiming to maximize the cumulative reward over time. The core difficulty lies in addressing the exploration-exploitation dilemma: the agent needs to explore different arms to learn their reward distributions while simultaneously exploiting the best-known arm to achieve immediate gains.

Balancing exploration and exploitation is crucial because exploration uncovers potentially better options, while exploitation ensures short-term performance. Overemphasizing exploration can waste resources on suboptimal choices, whereas overly exploiting known options risks missing higher rewards. This trade-off is central to the design of MAB algorithms.

MAB problems can be broadly categorized into two types: context-free bandits and contextual bandits. Context-free bandits have been extensively studied, with popular algorithms such as the $\epsilon$-greedy strategy~\cite{langford2007epoch-greedy} and the Upper Confidence Bound (UCB) algorithm~\cite{agrawal1995ucb}. These approaches assume that the rewards are solely determined by the arm selection, without considering additional information. In contrast, contextual bandits extend this framework by incorporating side information, or "context," to model the expected reward for each arm. Contextual bandits leverage the context as input features, enabling a more nuanced understanding of the reward function. For instance, algorithms like LinUCB~\cite{li2010personalized_recommendation} and Thompson Sampling for linear models~\cite{agrawal2013thompson} assume that the expected reward is a linear function of the context. However, in practice, this linearity assumption often fails for complex, non-linear environments.

To overcome this limitation, many works~\cite{ban2021multi} have integrated deep neural networks (DNNs) with contextual bandit frameworks, significantly enhancing their representation power. In our approach, we adopt a convolutional neural network (CNN) for contextual bandit algorithm to dynamically determine the optimal denoising timestep in video inference. Specifically, we model this problem as follows:
\begin{itemize}[leftmargin=*]
\item State Representation: The environment state is defined by the input video latent \(\bx_0^r\), capturing structural and perceptual details of the video.  
\item Agent and Policy: The agent is modeled using a CNN policy network, which employs a probabilistic policy \(\pi_\psi(t|\bx_0^r; \psi)\) to select the timestep \(t\).  
\item Action Selection: The action corresponds to the choice of a timestep \(t \in [0, T']\), representing the level of noise to add during the denoising process.  
\item Reward Signal: Feedback is provided in the form of a reward signal, defined as the negative mean squared error loss (\(\mathcal{L}_{MSE}\)) between the denoised output and the ground truth. This reward quantifies the quality of the denoising process for the chosen \(t\).  
\item Policy Update: The policy network updates its parameters \(\psi\) using the observed rewards, gradually learning to select the optimal \(t\) for different contexts.
\end{itemize}
By framing timestep selection as a contextual bandit problem, our method adaptively balances structural preservation and artifact correction during video inference, achieving robust and high-quality results across diverse scenarios.

\begin{figure*}[t]
    \centering
    \includegraphics[width=\linewidth]{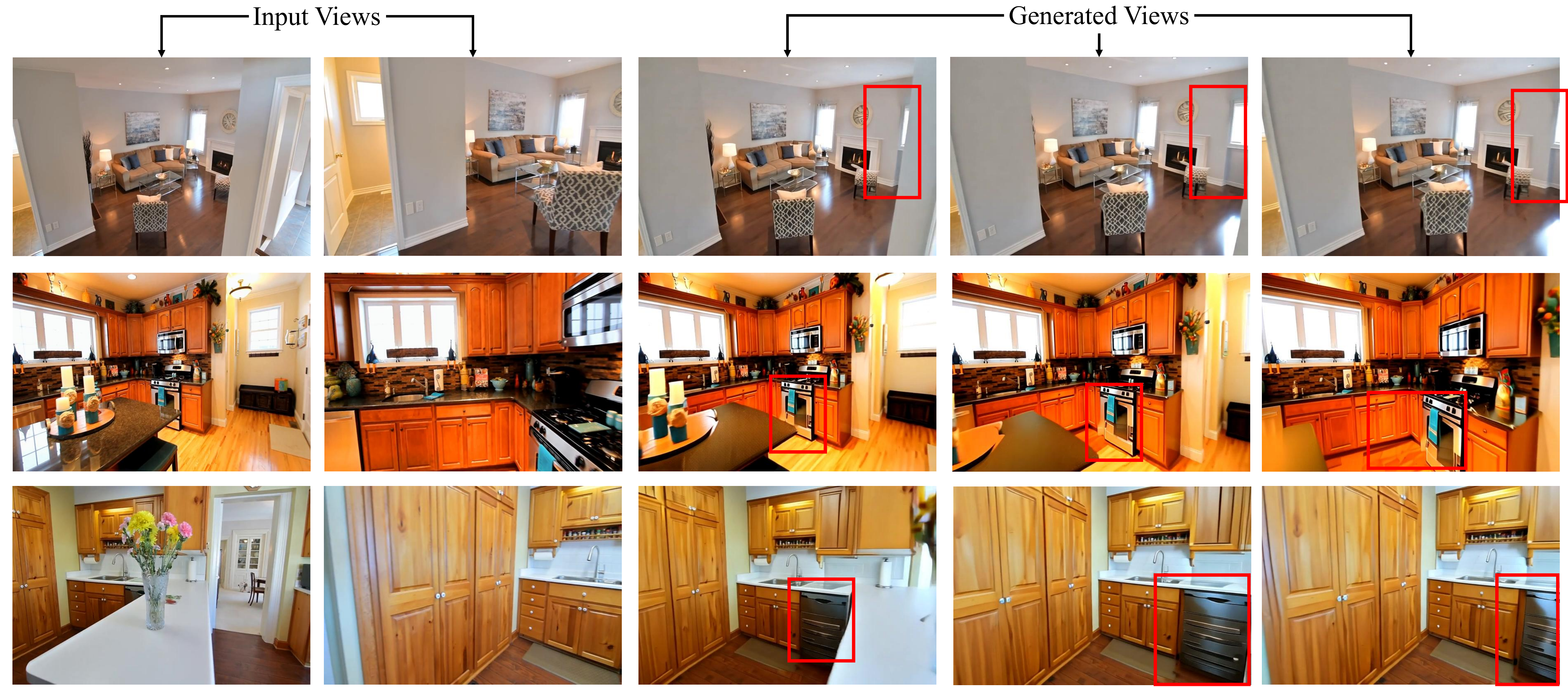}
    \caption{Visual results of the generative ability. We highlight the generated regions in the red boxes in the novel generated views.}
    \label{fig:generative_results}
\end{figure*}

\section{Additional Implementation Details}
We present the pseudo-code for 3D-aware distillation in Algorithm~\ref{tab:algorithm}. We also provide details for additional experiments.

\noindent\textbf{Datesets.} To further validate our strong generalizability, we test our method on the NeRF-LLFF~\citep{mildenhall2021nerf}, Sora~\cite{brooks2024sora}, and more challenging outdoor datasets Mip-NeRF 360~\citep{barron2022mipnerf360} and Tank-and-Temples dataset~\citep{knapitsch2017tanks}. For video-to-3D application, we also evaluate our method on the Mip-NeRF 360~\citep{barron2022mipnerf360} and Tank-and-Temples dataset~\citep{knapitsch2017tanks}.

\noindent\textbf{Video Metrics.}
We utilize VBench~\cite{huang2023vbench} to evaluate the performance of our model by comparing it against several state-of-the-art, open-source video frame interpolation models. VBench provides a comprehensive analysis of video generation quality by decomposing it into 16 distinct evaluation metrics, enabling a detailed and multi-faceted assessment of model performance. For our evaluation, we focus on key metrics such as Aesthetic Quality, Subject Consistency, and Background Consistency, which offer critical insights into the visual appeal and temporal coherence of the generated frames.

Aesthetic Quality measures the visual appeal of the generated video. Utilizing the LAION aesthetic predictor~\cite{laion2024aestheticpredictor}, it gauges the artistic and aesthetic value perceived by humans for each video frame. This score reflects various aesthetic dimensions, such as the layout, the richness and harmony of colors, photorealism, naturalness, and the artistic quality of the video frames. Subject Consistency assesses whether the appearance of a subject remains visually stable and coherent across all frames of a video. This metric is computed by evaluating the similarity of DINO~\cite{Caron_2021_ICCV} features extracted from consecutive frames. Background Consistency evaluates the temporal consistency of the background scenes by calculating CLIP~\cite{pmlr-v139-radford21a} feature similarity across frames.

To comprehensively evaluate the quality of the generated videos, we included additional metrics from VBench in the supplementary material, including Motion Smoothness, Dynamic Degree, and Imaging Quality. Motion Smoothness measures the fluidity of motion within the generated video, evaluating how well the movement follows realistic, natural trajectories. This metric assesses whether the video adheres to the physical laws governing motion in the real world. By utilizing motion priors in the video frame interpolation model~\cite{licvpr23amt}, it quantifies the temporal smoothness of the generated motions. Dynamic Degree is estimated by RAFT~\cite{teed2020raft} to indicate the temporal quality of generated videos. In our setting of still-scene video generation, an excessively high Dynamic Degree indicates unnecessary motion of objects within the scene, while an overly low Dynamic Degree suggests prolonged static periods interrupted by abrupt changes in certain frames. Both scenarios are undesirable outcomes for our task. Imaging Quality refers to the level of distortion present in the generated frames and is assessed using the MUSIQ~\cite{ke2021musiq} image quality predictor, which has been trained on the SPAQ~\cite{fang2020cvpr} dataset.

\noindent\textbf{Implementation Details for Video-to-3D Application.} 
For the video-to-3D application, we evaluate the 3D reconstruction performance of our method on the Mip-NeRF 360~\citep{barron2022mipnerf360} and Tanks-and-Temples datasets~\citep{knapitsch2017tanks}. Starting with two input images and corresponding camera poses estimated from DUSt3R~\cite{wang2024dust3r}, we first generate a continuous video sequence interpolating between the two frames. From this sequence, we extract intermediate frames by sampling every seventh frame, resulting in seven new views from novel perspectives. These sampled frames are then processed using InstantSplat~\cite{fan2024instantsplat} for Gaussian optimization-based 3D reconstruction from the generated novel views.

To assess the quality of our approach, we compare it against SparseNeRF~\citep{wang2023sparsenerf}, the original 3DGS~\citep{kerbl2023_3dgs}, and DNGaussian~\citep{li2024dngaussian}, with per-scene optimization serving as the benchmark. For quantitative evaluation, we report standard novel view synthesis (NVS) metrics, including PSNR, SSIM~\citep{wang2004ssim}, and LPIPS~\citep{zhang2018lpips}.

\begin{figure*}[t]
    \centering
    \includegraphics[width=\linewidth]{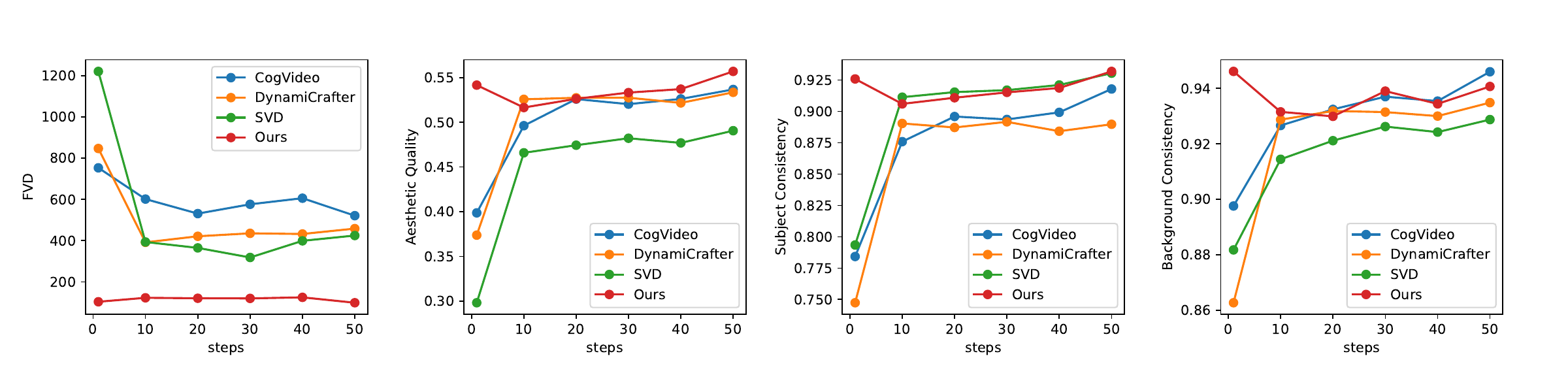}
    \caption{\textbf{Quantitative comparison across steps.} We evaluate the results of CogVideo, DynamiCrafter, Stable Video Diffusion (SVD), and VideoScene across 1, 10, 20, 30, 40, and 50 steps. VideoScene not only outperforms the other methods but also demonstrates remarkable consistency, with its 1-step results closely approximating its 50-step results, whereas other methods exhibit a significant decline in performance over fewer steps.}
    \label{fig:broken_line}
\end{figure*}

\begin{figure*}[t]
    \includegraphics[width=\linewidth]{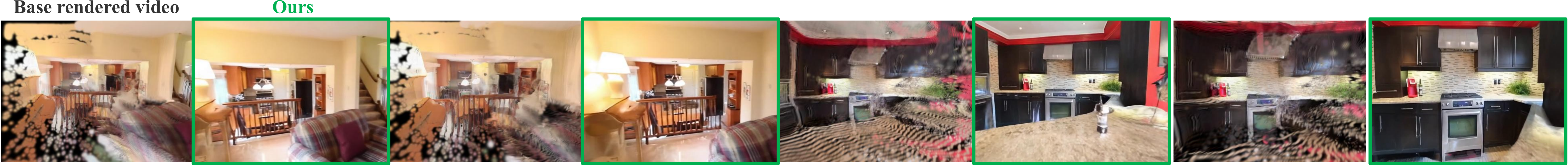}
    \caption{Comparisons with base renderings with severe artifacts.}
    \label{fig:base_rendered}
\end{figure*}

% 2.viewcrafter和ours
\begin{figure*}[t]
    \includegraphics[width=\linewidth]{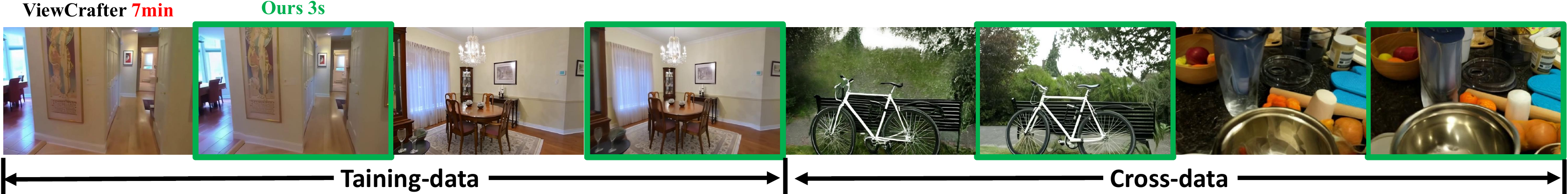}
    \caption{Comparisons with 3D-aware diffusion model ViewCrafter.}
    \label{fig:viewcrafter}
\end{figure*}

\begin{figure*}[h]
    \includegraphics[width=\linewidth]{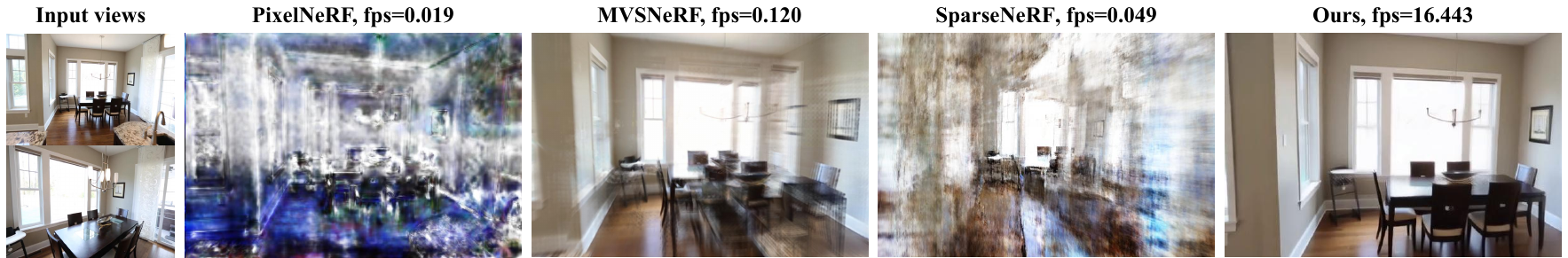}
    \caption{Comparisons with NeRF-based methods.}
    \label{fig:nerf}
\end{figure*}

\section{Additional Experiments and Analysis}
\subsection{More Visual Results}
\begin{figure*}[!t]
    \centering
    \includegraphics[width=0.85\linewidth]{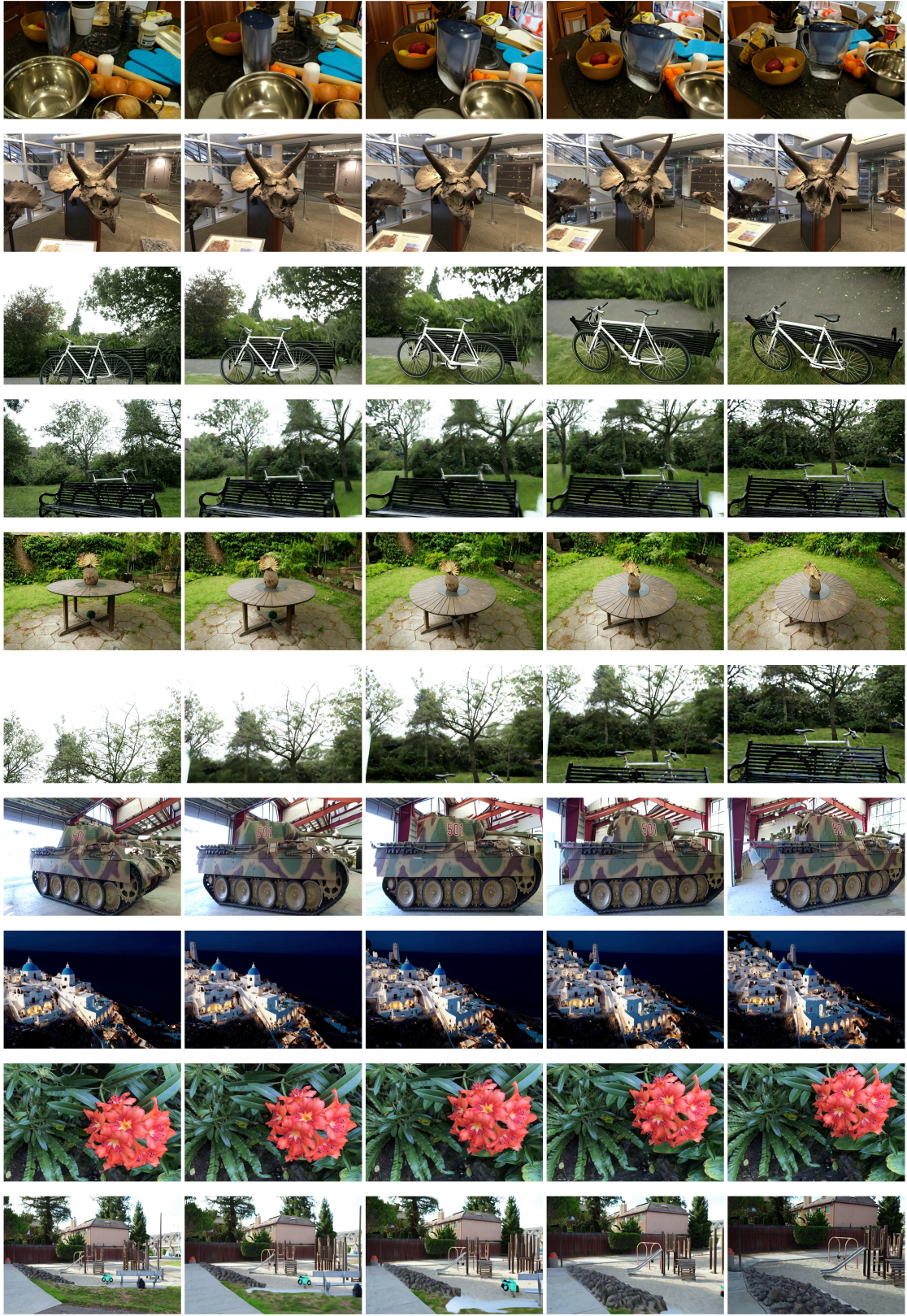}
    \caption{\textbf{Visual results of VideoScene.} We show visual results on NeRF-LLFF~\cite{mildenhall2021nerf}, Sora~\cite{brooks2024sora}, Mip-NeRF 360~\cite{barron2022mipnerf360}, and Tank-and-Temples dataset~\citep{knapitsch2017tanks} datasets. The first and last columns represent the input views, while the intermediate columns depict the generated views.}
    \label{fig:frames}
\end{figure*}

We present additional visual results of our VideoScene framework in Fig.~\ref{fig:frames}, showcasing its performance across diverse datasets, including NeRF-LLFF~\citep{mildenhall2021nerf}, Sora~\citep{brooks2024sora}, Mip-NeRF 360~\citep{barron2022mipnerf360}, and Tanks-and-Temples~\citep{knapitsch2017tanks}. These examples highlight the strong generalization capability of our method, effectively adapting to novel and out-of-distribution scenarios, whether indoor or outdoor.

We also provide additional visual results in Fig.~\ref{fig:generative_results} to further illustrate the generative capability of our model. When the input consists of two images with significantly different viewpoints, the intermediate regions often lack direct coverage by either input image. In such cases, a model must rely on its generative ability to synthesize these unseen areas. As highlighted by the red boxes in Fig.~\ref{fig:generative_results}, VideoScene successfully generates novel content for these unseen regions. This demonstrates not only the strong generative capacity of our model but also its ability to generalize effectively while maintaining high fidelity in reconstructing previously unobserved areas.

\begin{table}[!t]
    \begin{center}
    \caption{Quantitative comparison on Mip-Nerf 360 and Tank-and-Temples datasets. We report the quantitative metrics with two input views for each scene.}
    
    \label{tab:3d}
    \resizebox{1\linewidth}{!}{
    \begin{tabular}{lccc}
    \toprule

    Method & PSNR$\uparrow$ & SSIM$\uparrow$ & LPIPS$\downarrow$ \\
    
    \midrule
    \textbf{Mip-NeRF 360} \\
    3DGS & 10.36 & 0.108 & 0.776 \\
    SparseNeRF & 11.47  & 0.190 & 0.716 \\ 
    
    DNGaussian & 10.81  & 0.133 & 0.727 \\

    InstantSplat & 11.77  & 0.171 & 0.715 \\ 

    Ours & \textbf{13.37} & \textbf{0.283} & \textbf{0.550} \\ 
    
    \midrule
    \textbf{Tank and Temples} \\
    3DGS & 9.57  & 0.108 & 0.779 \\
    SparseNeRF &  9.23  & 0.191 & 0.632 \\ 
    
    DNGaussian & 10.23  & 0.156 & 0.643 \\ 

    InstantSplat & 10.98  & 0.381 & 0.619 \\ 

    Ours & \textbf{14.28} & \textbf{0.394} & \textbf{0.564} \\ 
    \bottomrule
    \end{tabular}
    }
    \end{center}
\end{table}

\begin{figure}[t]
    \centering
    \includegraphics[width=\linewidth]{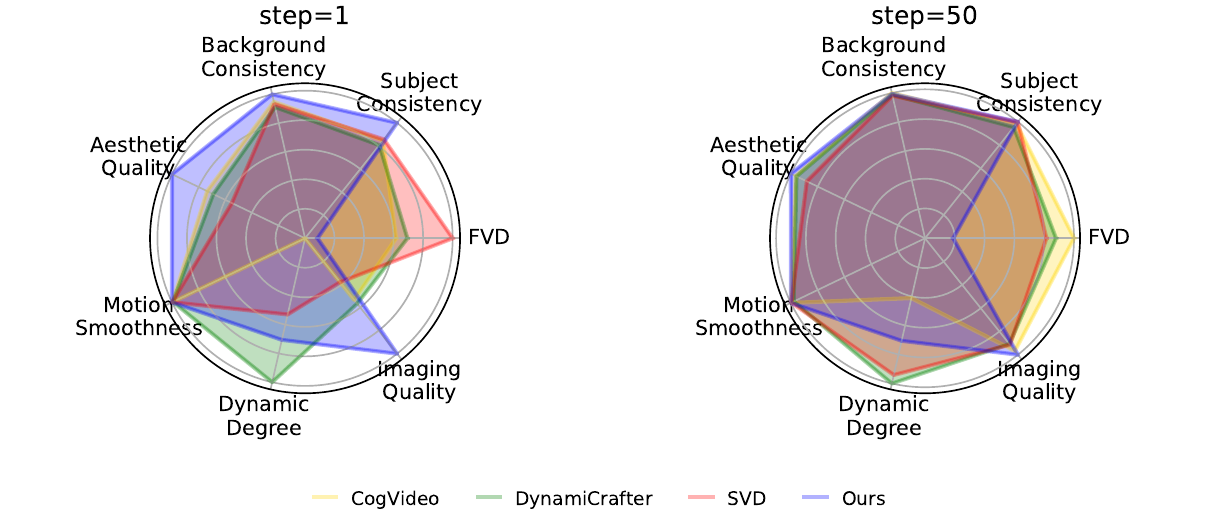}

    \caption{\textbf{Quantitative comparison across additional dimensions.} We further evaluate the 1-step and 50-step results by incorporating additional dimensions from the VBench metrics.}
    \label{fig:radar}
\end{figure}

\subsection{More Quantitative Comparison Results}
We provide comprehensive quantitative comparisons with baseline methods in Fig.~\ref{fig:broken_line}, \ref{fig:radar}. In Fig.~\ref{fig:broken_line}, we evaluate the performance of CogVideo~\cite{yang2024cogvideox}, DynamiCrafter~\cite{xing2025dynamicrafter}, Stable Video Diffusion (SVD)~\cite{blattmann2023svd}, and our VideoScene across different inference steps. The results demonstrate that VideoScene not only surpasses other methods in generation quality but also achieves results comparable to their 50-step outputs in just one step. In contrast, the one-step outputs of other methods fall significantly behind their 50-step counterparts, highlighting the efficiency and effectiveness of our approach.

In Fig.~\ref{fig:radar}, we further evaluate our method across multiple dimensions using metrics from VBench~\cite{huang2023vbench}, providing a more systematic and holistic validation of our generative quality. Notably, the Dynamic Degree metric assesses both the dynamic motion of individual objects in the scene and overall camera motion. Our method carefully balances these aspects, preserving consistent camera motion while minimizing unstable object movements, resulting in a well-balanced intermediate Dynamic Degree value. In comparison, DynamiCrafter exhibits higher values due to its inability to maintain relative object stability, leading to excessive motion. Conversely, CogVideo shows lower values, as it often produces videos with prolonged static periods interrupted by abrupt transitions, particularly between the first and second halves. These observations underscore the robustness and balanced performance of our approach.

\subsection{More Qualitative Comparison Results}
In Fig.~\ref{fig:base_rendered}, we compare our VideoScene with MVSplat base renderings to show its effectiveness. In Fig.~\ref{fig:viewcrafter}, we compare with another 3D-aware diffusion model~\cite{yu2024viewcrafter}, and in Fig.~\ref{fig:nerf}, we show more visual comparison with NeRF-based methods~\cite{yu2021pixelnerf,chen2021mvsnerf,wang2023sparsenerf}. 

\subsection{Video-to-3D Applications}
\begin{figure*}[!t]
    \centering
    \includegraphics[width=\linewidth]{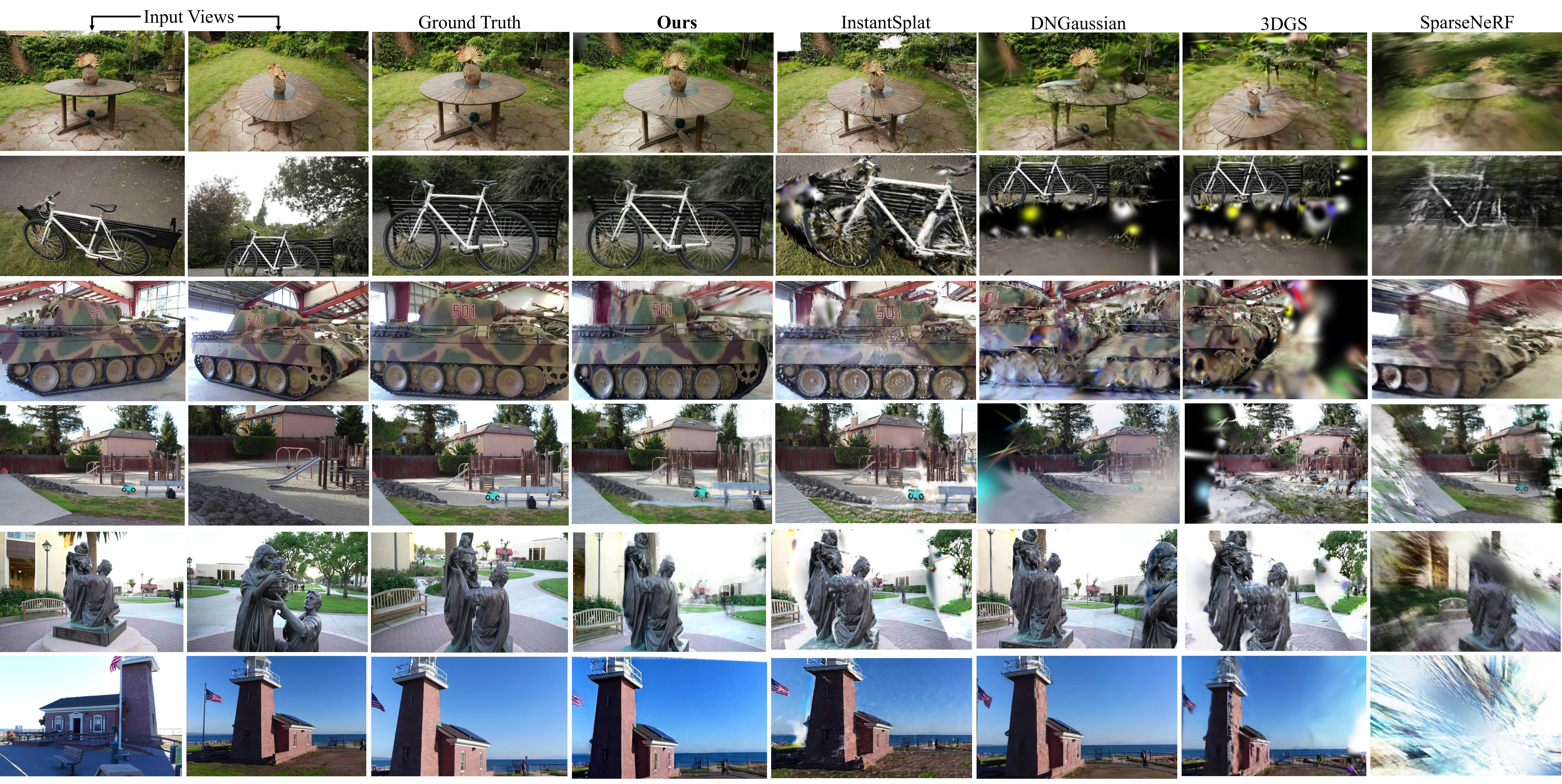}
    \caption{Qualitative comparison on Mip-Nerf 360 and Tank-and-Temples. With two sparse views as input, our method achieves much better reconstruction quality compared with baselines.}
    \label{fig:3d}
\end{figure*}

We evaluate the geometric consistency of our generated frames to assess their suitability for downstream tasks such as 3D reconstruction. For this purpose, we utilize InstantSplat~\cite{fan2024instantsplat}, a 3D Gaussian Splatting (3DGS) method built on DUSt3R~\cite{wang2024dust3r}, which generates Gaussian splats from sparse, unposed images. Using this approach, we use VideoScene to generate video frames from given two input views and optimize the generated frames for 3D Gaussian representations. We also compare our method against existing per-scene optimization techniques, including Instantsplat~\cite{fan2024instantsplat}, DNGaussian~\cite{li2024dngaussian}, 3DGS~\cite{kerbl2023_3dgs}, and SparseNeRF~\cite{wang2023sparsenerf}. The results, presented in Tab.~\ref{tab:3d} and Fig.~\ref{fig:3d}, demonstrate that our approach effectively preserves the geometric structure of the scene, avoiding issues such as the multi-face problem. Furthermore, our method exhibits strong generative capabilities, reconstructing regions beyond the coverage of input views.

\begin{table}[t]
    \centering
    \small
    \caption{User study on the layout stability, smoothness, visual realism, and overall quality score in a user study, rated on a range of 1-10, with higher scores indicating better performance.}
    \label{tab:user_study}
    \resizebox{\linewidth}{!}{
    \begin{tabular}{lcccc}
        \toprule
        Method  & Layout Stability & Smoothness & Visual Realism & Overall Quality \\
        \midrule
        Stable Video Diffusion~\cite{blattmann2023svd} & 6.48 & 7.29 & 6.75 & 7.13 \\
        DynamiCrafter~\cite{xing2025dynamicrafter} & 7.02 & 7.01 & 6.02 & 6.68 \\
        CogVideoX~\cite{yang2024cogvideox} & 7.83 & 7.53 & 7.33 & 7.50 \\
        \midrule
        \textbf{Ours} & \textbf{8.39} & \textbf{8.91} & \textbf{9.52} & \textbf{8.82} \\
        \bottomrule
    \end{tabular}}
\end{table}

\begin{figure*}[h]
    \includegraphics[width=\linewidth]{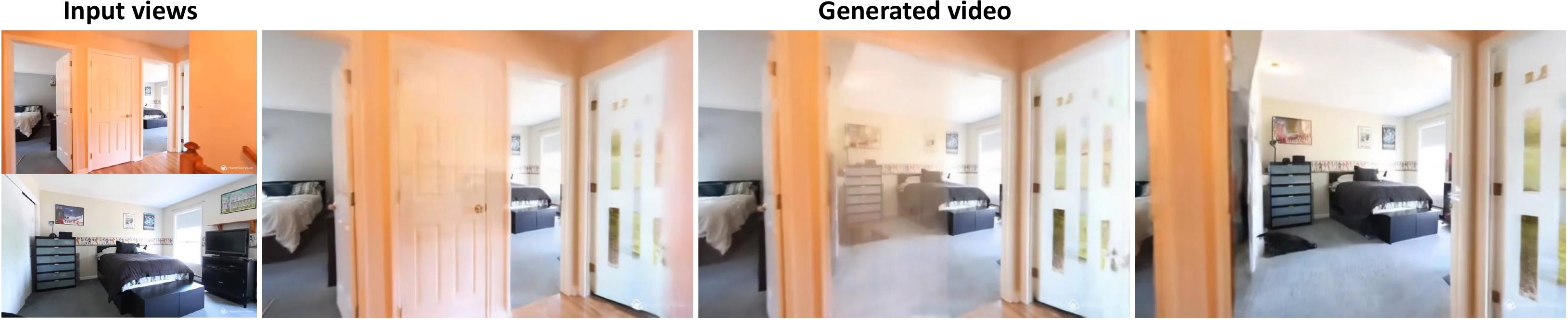}
    \caption{Fail case of passing directly through the closed door.}
    \label{fig:fail_case}
\end{figure*}

\subsection{User Study}
For the user study, we show each volunteer five samples of generated video using a random method. They can rate in four aspects: (1) \textit{layout stability.} Users assess whether the scene layout in the video is spatially coherent and consistent. (2) \textit{smoothness.} Users observe whether the frame rate is stable, whether actions are smooth, and whether there are any stuttering or frame-skipping issues. (3) \textit{visual realism.} Users rate the similarity between the generated video and a real video. (4) \textit{overall quality}. All aspects are on a scale of 1-10, with higher
scores indicating better performance. We collect results from 30 volunteers shown in Table~\ref{tab:user_study}. We find users significantly prefer our method over these aspects.

\subsection{Failure Case}
Significant semantic disparities between input views lead to failure cases (see Fig.~\ref{fig:fail_case}). The generated video passes directly through the closed door rather than navigating around it to enter the room.

\section{More Discussions}
\subsection{Discussion of Empirical Runtime}
We provide the runtime comparison in Tab~\ref{tab:runtime}. DynamiCrafter is efficient due to smaller model size and lower frame rates and resolutions, but ours is still much faster.

\begin{table}[!h]
    \centering
    \caption{Empirical runtime comparisons.}
    \resizebox{1\linewidth}{!}{
    \begin{tabular}{lccccc}
        \toprule
        Method  & SVD & DynamiCrafter & CogVideoX-5B & ViewCrafter & VideoScene (\textbf{Ours}) \\
        \midrule
        Runtime (s) & 933.89 & 21.14 & 179.45 & 206.13 & \textbf{2.98} \\
        Frames & 25 & 16 & \textbf{49} & 25 & \textbf{49} \\
        \bottomrule
    \end{tabular}}
    \label{tab:runtime}
\end{table}

\subsection{Discussion of limited computational resources}

Tab.~\ref{tab:memory} presents the comparison of memory costs on a single A100. The primary consumer of computational resources is the video diffusion model itself, which is inherently unavoidable. Leap flow distillation, as a strategy for video training, incurs a computational cost comparable to that of video diffusion training, without introducing significant additional overhead.

\begin{table}[!h]
    \centering
    \caption{Comparison on memory costs.}
    \resizebox{1\linewidth}{!}{
    \begin{tabular}{lcccccc}
        \toprule
         Description & Video Backbone (CogVideoX) & Leap Flow Distillation & DDPNet  & Total \\
        \midrule
        Training Cost & $\sim 66$ GB & $\sim10$ GB & $\sim0.02$ GB & $\sim 76$ GB\\
        \bottomrule
    \end{tabular}}
    \label{tab:memory}
\end{table}

\end{document}